\documentclass[lettersize,journal]{IEEEtran}
\usepackage{amsmath,amsfonts}
\usepackage{algorithmic}
\usepackage{algorithm}
\usepackage{array}
\usepackage[caption=false,font=normalsize,labelfont=sf,textfont=sf]{subfig}
\usepackage{textcomp}
\usepackage{stfloats}
\usepackage{url}
\usepackage{verbatim}
\usepackage{graphicx}
\usepackage{cite}
\usepackage{soul}
\usepackage{color}
\usepackage{gensymb}
\usepackage{booktabs}
\usepackage{gensymb}
\usepackage{xspace}
\usepackage[table]{xcolor}
\usepackage[pagebackref,breaklinks,colorlinks,citecolor=teal]{hyperref}
\usepackage[capitalise]{cleveref}

\newcommand{\synpain}{Syn\textsc{Pain}\xspace}

\begin{document}

\title{\synpain: A Synthetic Dataset of Pain and Non-Pain Facial Expressions }

\author{Babak Taati,
Muhammad Muzammil, Yasamin Zarghami, Abhishek Moturu, \\Amirhossein Kazerouni, Hailey Reimer, Alex Mihailidis, Thomas Hadjistavropoulos
\thanks{Babak Taati is with the KITE Research Institute, Toronto Rehabilitation Institute, University Health Network; the Department of Computer Science, University of Toronto; the Institute of Biomedical Engineering, University of Toronto; and the Vector Institute; Toronto, Canada (email: Babak.Taati@uhn.ca).}
\thanks{Muhammad Muzammil is with the Department of Computer Science, University of Toronto; Toronto, Canada.}
\thanks{Yasamin Zarghami, Abhishek Moturu, and Amirhossein Kazerouni are with the KITE Research Institute, Toronto Rehabilitation Institute, University Health Network; the Department of Computer Science, University of Toronto; and the Vector Institute; Toronto, Canada.}
\thanks{Alex Mihailidis is with the KITE Research Institute, Toronto Rehabilitation Institute, University Health Network; the Department of Occupational Science and Occupational Therapy, University of Toronto; and the Institute of Biomedical Engineering, University of Toronto; Toronto, Canada.}
\thanks{Hailey Reimer and Thomas Hadjistavropoulos are with the University of Regina; Regina, Canada.}}

%

\maketitle

\begin{abstract}

Accurate pain assessment in patients with limited ability to communicate, such as older adults with severe dementia, represents a critical healthcare challenge. Robust automated systems of pain behavior detection may facilitate such assessments. Existing pain detection datasets, however, suffer from limited ethnic/racial diversity, privacy constraints, and underrepresentation of older adults who are the primary target population for clinical deployment. We present \synpain, a large-scale synthetic dataset containing 10,710 facial expression images (5,355 neutral/expressive pairs) across five ethnicities/races, representing two age groups (young:~20--35, old:~{75--93}), and two genders. Using commercial generative AI tools, we created demographically balanced synthetic identities with clinically meaningful pain expressions. Our validation demonstrates that synthetic pain expressions exhibit expected pain patterns, scoring significantly higher than neutral and non-pain expressions using clinically validated pain assessment tools based on facial action unit analysis. We experimentally demonstrate \synpain's utility in identifying algorithmic bias in existing pain detection models. Through comprehensive bias evaluation, we reveal substantial performance disparities across demographics characteristics. These performance disparities were previously undetectable with smaller, less diverse datasets. Furthermore, we demonstrate that age-matched synthetic data augmentation improves pain detection performance on real clinical data, achieving a {2.4 percentage point improvement} in average precision. \synpain addresses critical gaps in pain assessment research by providing the first publicly available, demographically diverse synthetic dataset specifically designed for older adult pain detection, while establishing a framework for measuring and mitigating algorithmic bias. {The dataset, code, and trained models is available at \href{https://mmzml.github.io/SynPAIN}{\synpain}.}

\end{abstract}

\begin{IEEEkeywords}
Synthetic Data, Pain Detection, Facial Expression Recognition, Algorithmic Bias, Generative AI, Data Augmentation.
\end{IEEEkeywords}

\section{Introduction}

\IEEEPARstart{A}{ccurate} pain assessment, particularly for patients who cannot self-report their experience due to cognitive and linguistic impairments, represents a critical healthcare challenge. While automated pain assessment systems require large, diverse datasets to train robust AI models, collecting and labeling facial expression data is time-consuming, costly, and often limited by privacy concerns. This paper introduces \synpain~\cite{SP3/WCXMAP_2025}, a synthetic dataset of pain and non-pain facial expressions to overcome traditional data collection limitations and advance automated pain detection through demographically diverse training data.

\subsection{Pain in Older Adults with Dementia}

For older adults with moderate to severe dementia, pain assessment relies heavily on nonverbal cues, as cognitive impairment often interferes with the ability to self-report pain~\cite{hadjistavropoulos2005assessing}. Under-assessment and under-treatment of pain in this population are well documented and can have devastating consequences for quality of life~\cite{achterberg2021chronic}. Untreated pain is also a potential cause of agitation and aggression in dementia care facilities~\cite{cipher2006behavioral}.

Although frequent pain assessment is critically important and supported by expert consensus~\cite{hadjistavropoulos2007interdisciplinary,herr2011pain,hadjistavropoulos2014pain,herr2019pain}, regular in-person evaluations are costly and often not implemented in long-term care settings~\cite{weissman1999pain,gagnon2013development,guliani2021pain,pringle2021pain}. Automated monitoring systems offer a promising solution by enabling continuous, objective pain assessment and timely notification of staff when intervention is needed~\cite{kunz2017problems}.

\subsection{Challenges in Data Collection}

Large facial image or video datasets are needed to train AI models that automatically analyze facial expressions, such as detecting pain in non-verbal patients. Collecting and labeling facial image/video data is time-consuming and extremely costly, and when data must be collected from patient populations, it can be burdensome for them, their families, and caregivers. Privacy and confidentiality concerns further limit the possibility of sharing or making video-recorded behaviors publicly available. Additionally, demographic factors such as participant ethnicity, age, or gender can affect research results; lack of diverse training data leads to algorithmic bias~\cite{buolamwini2018gender}. 

The development of robust automated pain detection systems for older adults with dementia faces additional challenges. Long-term care facilities are often difficult to access for research, as most are understaffed, under-resourced, and not incentivized to participate in research. Despite strong interest from family members and residents in contributing to clinical research~\cite{avent2013establishing}, ethics board requirements make it difficult to identify and obtain consent from potential participants, often necessitating proxy consent from family members or legal guardians. Ethics restrictions frequently prohibit the reuse of collected videos in future studies, forcing researchers to collect new data even for similar studies. These limitations highlight the need for datasets that are more age-diverse, easily shareable, and comprehensively annotated for age, gender, and ethnicity. In this work, we explore the utility of generative AI models in resolving these issues by presenting a large and diverse dataset of synthetic identities and facial expressions to investigate algorithmic bias and augment pain detection model training.

\section{Previous Work}

\subsection{Clinically Validated Methods of Pain Assessment}

Health psychology researchers have developed two clinically validated metrics for assessing pain in older adults with dementia. The first is based on the Facial Action Coding System (FACS), an anatomically grounded taxonomy of facial movements that deconstructs expressions into distinct muscles or groups of muscles known as Action Units (AUs)~\cite{ekman1978facial}. Prkachin and Solomon validated a scoring approach using FACS that focuses exclusively on AUs consistently associated with pain~\cite{prkachin2008structure}. Their Prkachin and Solomon Pain Index (PSPI) is a score in the [0,16] range calculated as shown in \cref{eqn:PSPI}, with each contributing AU described in \cref{tab:AUDescriptions}.
\begin{equation}
\label{eqn:PSPI}
\small
PSPI = AU_4 + max(AU_6, AU_7) + max(AU_9, AU_{10}) + AU_{43}
\end{equation}

Another clinically validated system is the Pain Assessment Checklist for Seniors with Limited Ability to Communicate-II (PACSLAC-II)~\cite{chan2014evidence}. Accurately coding facial AUs requires extensive training and is time-consuming. In contrast, the PACSLAC-II offers the advantage of requiring less training and faster coding, as it relies on observable behavioral indicators rather than detailed facial muscle analysis.

\begin{table*}[!ht]
\caption{Description facial action units (AUs) used in the Prkachin and Solomon Pain Index (PSPI).}
\label{tab:AUDescriptions}
\centering
\begin{tabular}{l|c|l|l}
\textbf{Action Unit} & \textbf{Range}      & \textbf{Description}      & \textbf{Facial Muscle(s) }                                     \\ \midrule
AU4         & {[}0, 5{]} & Brow Lowerer     & depressor glabella, depressor supercilii, corrugator supercilii \\
AU6         & {[}0, 5{]} & Cheek Raiser     & orbicularis oculi, pars orbitalis                     \\
AU7         & {[}0, 5{]} & Lid Tightener    & orbicularis oculi, pars palpebralis                   \\
AU9         & {[}0, 5{]} & Nose Wrinkler    & levator labii superioris alaquae nasi                 \\
AU10        & {[}0, 5{]} & Upper Lip Raiser & levator labii superioris, caput infraorbitalis        \\
AU43        & \{0, 1\} & Eyes Closed      & relaxation of levator palpebrae superioris           
\end{tabular}
\end{table*}

\subsection{Existing Public Datasets}

The UNBC-McMaster Shoulder Pain Expression Archive Database~\cite{lucey2011painful} contains 200 video sequences (48,398 frames) of pain and non-pain facial expressions from 25 participants with chronic shoulder pain, while the BioVid Heat Pain Database~\cite{walter2013biovid} includes multimodal data from 90 healthy adults aged 20--65 exposed to thermal pain stimuli. Although widely used for benchmarking pain detection algorithms, these datasets lack representation of older adults and do not account for age-related facial morphology (e.g., pronounced wrinkles), limiting their utility for geriatric and dementia care. A systematic review underscores this demographic gap, noting that most publicly available pain datasets prioritize younger cohorts~\cite{gkikas2023automatic}.

Age-related facial changes, such as wrinkles, reduced collagen, and loss of elastic fibers~\cite{chung2001modulation,yaar2002fifty,quan2015role}, alter the visibility and dynamics of pain-related facial AUs. Comorbidities common in older adults, including post-stroke facial paresis~\cite{konecny2011facial,volk2019facial}, further introduce asymmetries or atypical expressions not captured in existing datasets~\cite{taati2019algorithmic,asgarian2019limitations,bandini2020new,guarin2020toward}. Consequently, computer vision models trained on younger populations often struggle to generalize to older adults~\cite{rezaei2020unobtrusive}.

FACS datasets such as DISFA~\cite{mavadati2013disfa} (130,798 frames from 27 participants) and BP4D/BP4D+~\cite{zhang2014bp4d,zhang2016multimodal} enable automatic detection of AU combinations~\cite{ning2024representation,yuan2024auformer,yang2019facs3d,onal2019d} for PSPI derivation. However, these resources also predominantly represent younger, healthy populations: Cohn-Kanade (CK)~\cite{kanade2000comprehensive} includes participants aged 18--30, CK+~\cite{lucey2010extended} expands to 18--50, Bosphorus~\cite{savran2008bosphorus} features ages 25--35, and BP4D focuses on 18--29. While BP4D+ extends to 18--66 years, older adults---particularly those in long-term care, where the average age exceeds 83~\cite{egbujie2024trajectories}---remain underrepresented.

This lack of age diversity in pain and FACS datasets impedes the development of reliable automated pain assessment systems for geriatric populations. Addressing this gap requires clinically representative datasets that reflect the anatomical and physiological diversity of aging faces, especially in vulnerable groups such as older adults with dementia.

\subsection{Automated Pain Assessment}

The vast majority of research on vision-based pain assessment (e.g.~\cite{ashraf2007painful, xu2019pain, tavakolian2019spatiotemporal, rau2024video, fiorentini2022fully,benavent2023comprehensive}) relies on publicly available datasets, particularly the UNBC-McMaster and BioVid Heat Pain datasets~\cite{werner2019automatic,gkikas2023automatic}. Automated pain assessment systems have also been developed for populations unable to self-report, including infants~\cite{brahnam2006machine,zamzmi2019convolutional,zamzmi2019comprehensive,brahnam2023neonatal}, partially sedated patients~\cite{kobayashi2021semi,zarghami2023pain}, and older adults with dementia~\cite{rezaei2020unobtrusive,stopyn2025real}, often using locally collected, non-public data. 

The current state-of-the-art (SOTA) model for detecting facial expressions of pain in older adults is the Pairwise with Contrastive Training (PwCT) model~\cite{rezaei2020unobtrusive}, which is trained on a combination of the public UNBC-McMaster dataset and the non-publicly available (due to ethical considerations) University of Regina (UofR) dataset. The UofR dataset comprises video recordings from 102 older adult participants, including individuals both with and without dementia, captured during baseline and pain-inducing phases. Trained evaluators manually annotated videos of 95 individuals from this dataset (74 women) using both PSPI and PACSLAC-II pain assessment frameworks. Among these 95 older adult participants, 47 were community-dwelling individuals with normal cognitive function (average age: 75.5\,$\pm$\,6.1 years), while the remaining 48 individuals (average age: 82.5\,$\pm$\,9.2 years) had severe dementia and were residents of long-term care facilities.

The PwCT model achieves robust performance through two key innovations: personalized neutral baselines and contrastive representation learning~\cite{rezaei2020unobtrusive}. By comparing test-time facial expressions to individualized neutral references, the model reduces sensitivity to age-related facial idiosyncrasies such as wrinkles and asymmetry, while maintaining responsiveness to pain-specific action unit dynamics. Contrastive training further enhances cross-dataset performance~\cite{rezaei2020unobtrusive}. This model has been externally validated \emph{in vivo}~\cite{stopyn2025real} with 65 cognitively healthy older adults (age: 71.8\,$\pm$\,5.8) and is currently being evaluated \emph{in situ} in four nursing homes in Saskatchewan, Canada. However, because the model was trained using a dataset that is not publicly available, there remains a critical need for a publicly available, shareable dataset of older adults with and without pain to advance research and support reproducibility in this field.

\begin{figure*}[t]
    \centering
    \includegraphics[width=.24\textwidth]{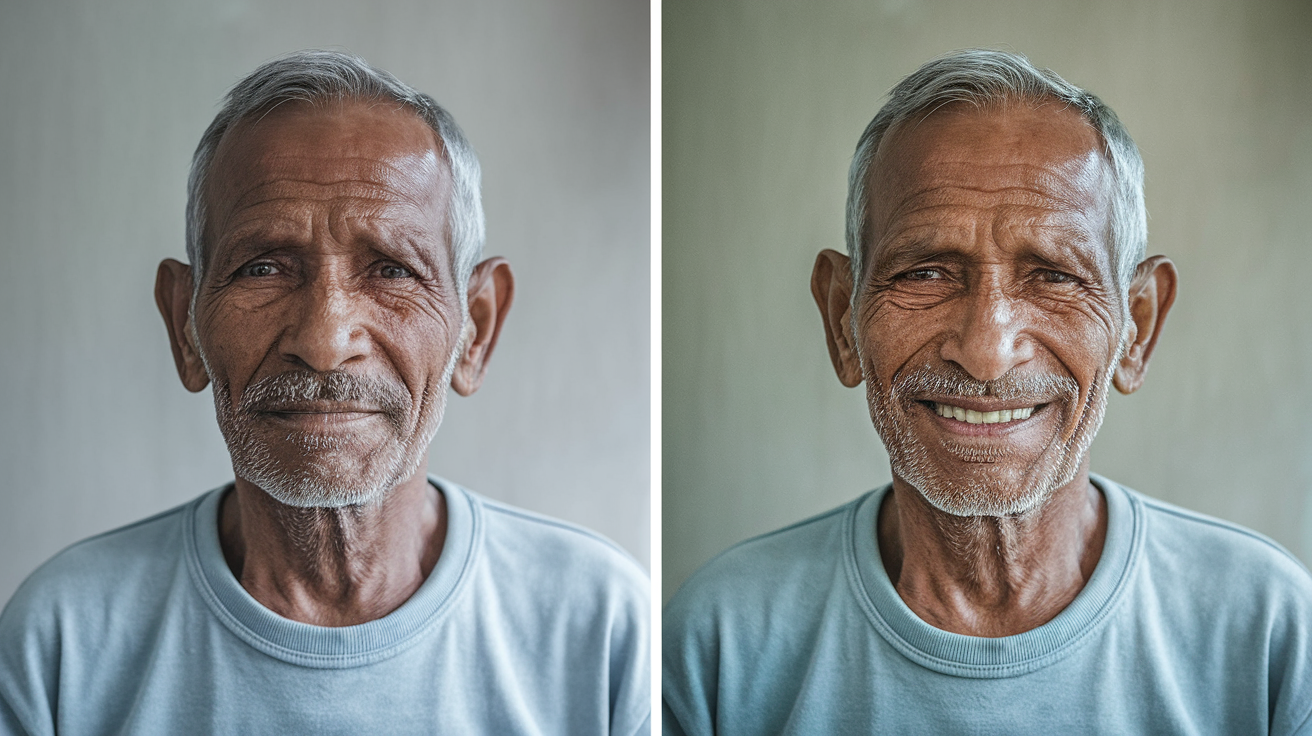}
    \includegraphics[width=.24\textwidth]{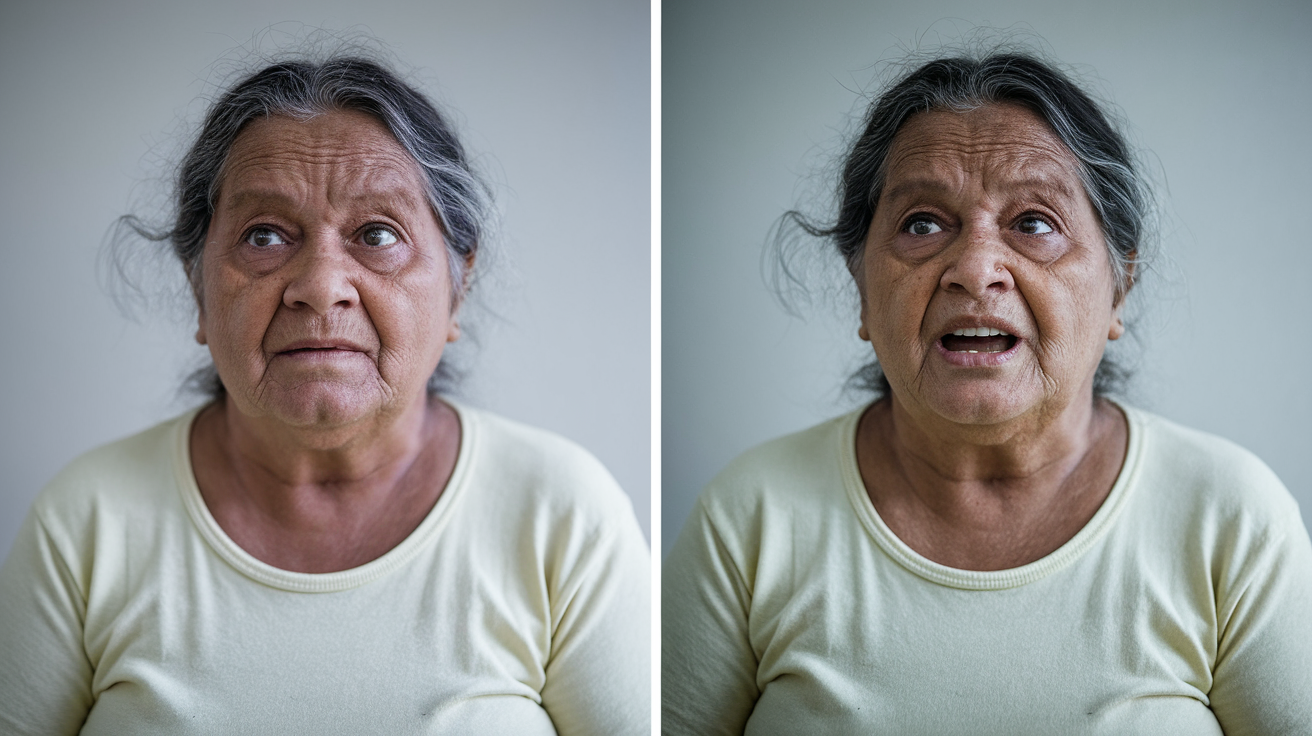}
    \includegraphics[width=.24\textwidth]{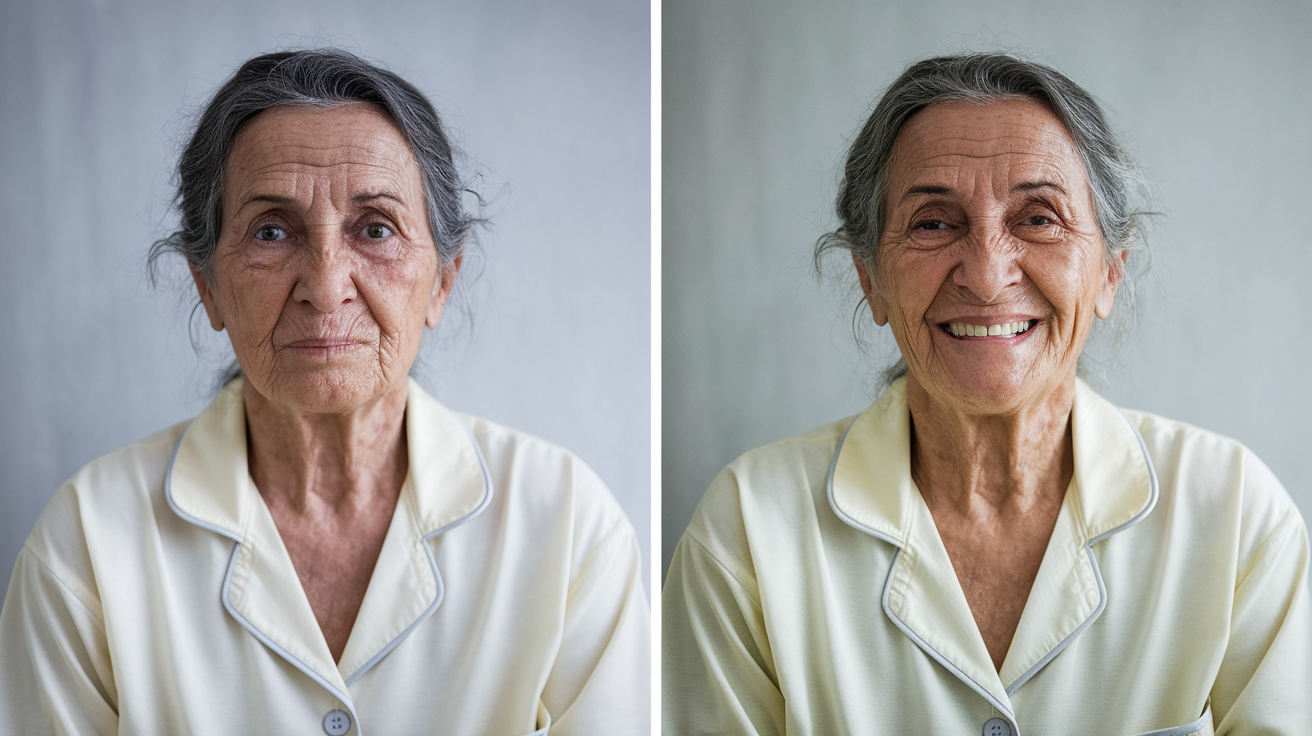}
    \includegraphics[width=.24\textwidth]
    {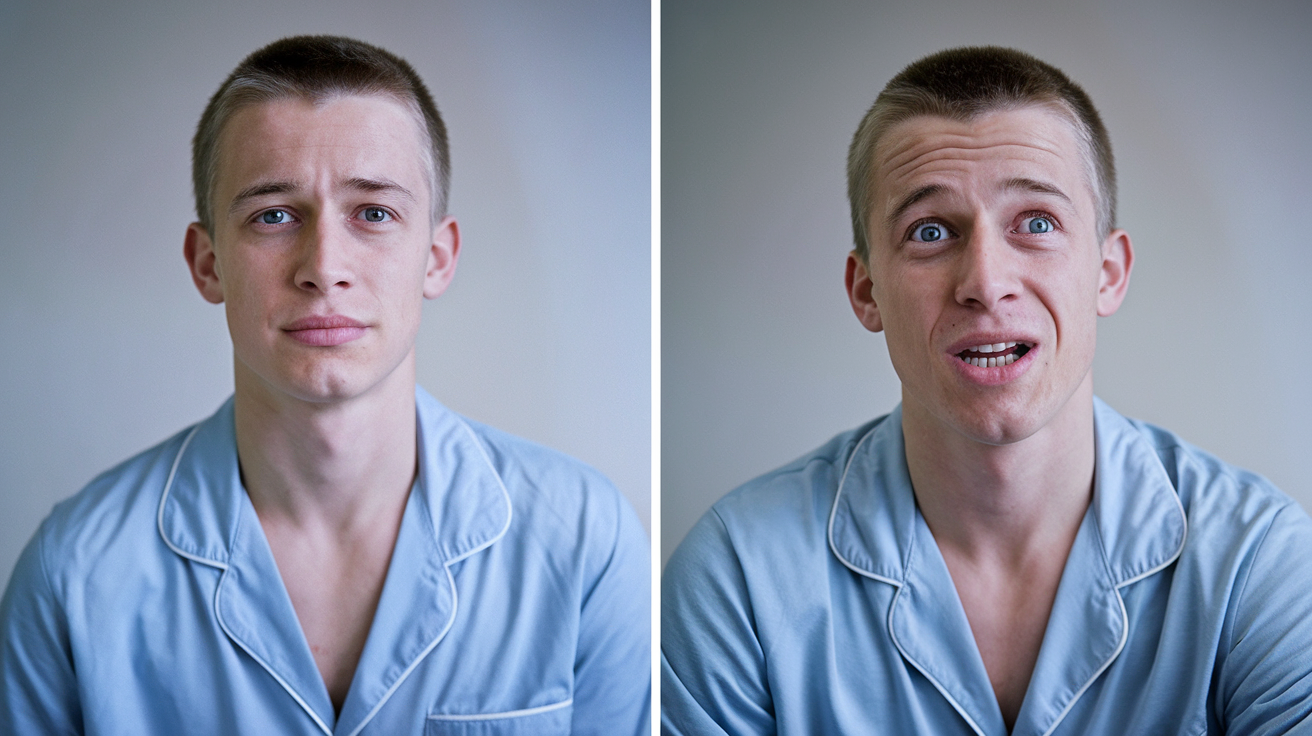}\\[1mm]
    \includegraphics[width=.24\textwidth]
    {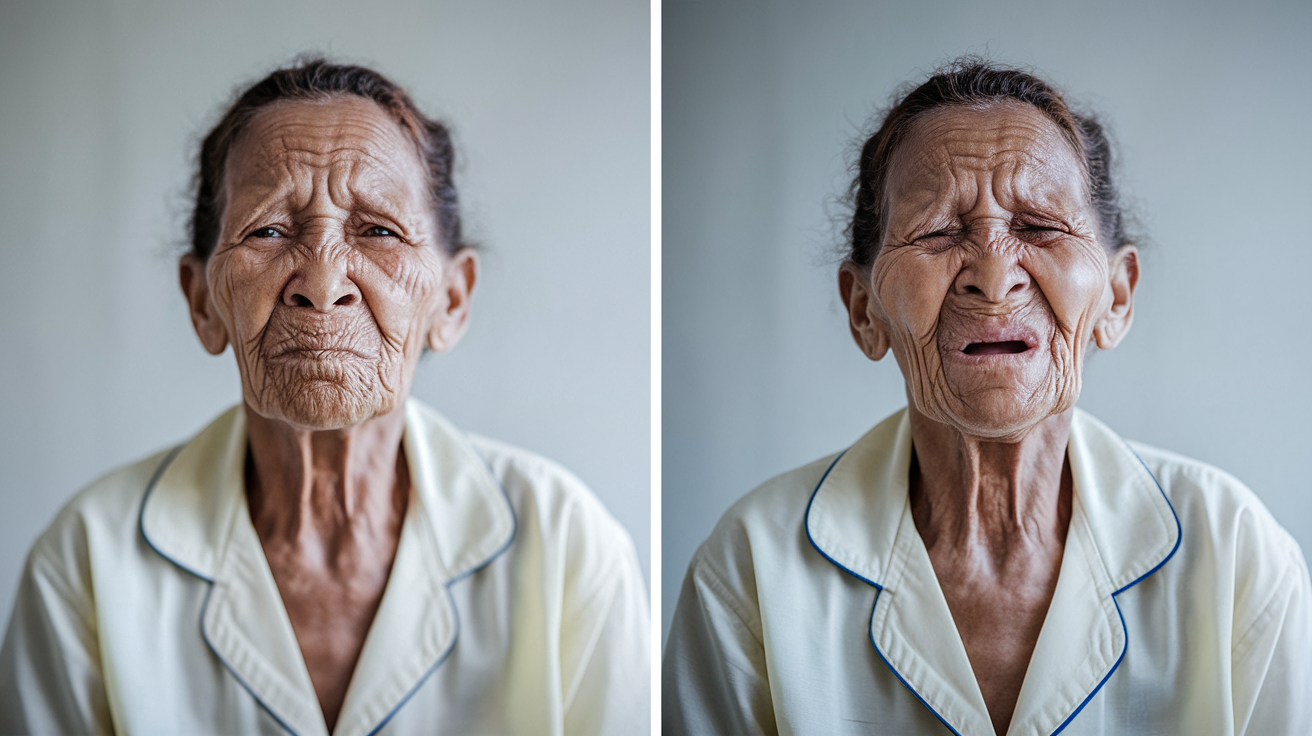}
    \includegraphics[width=.24\textwidth]
    {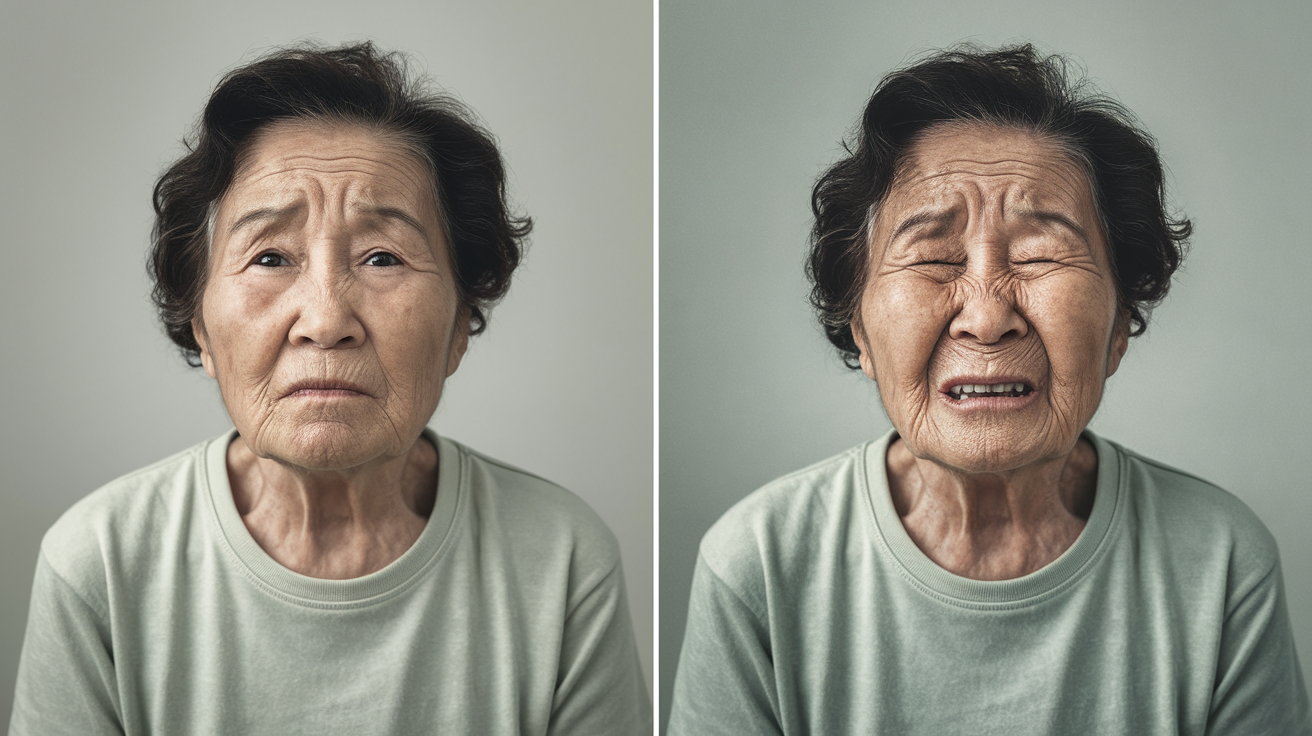}
    \includegraphics[width=.24\textwidth]
    {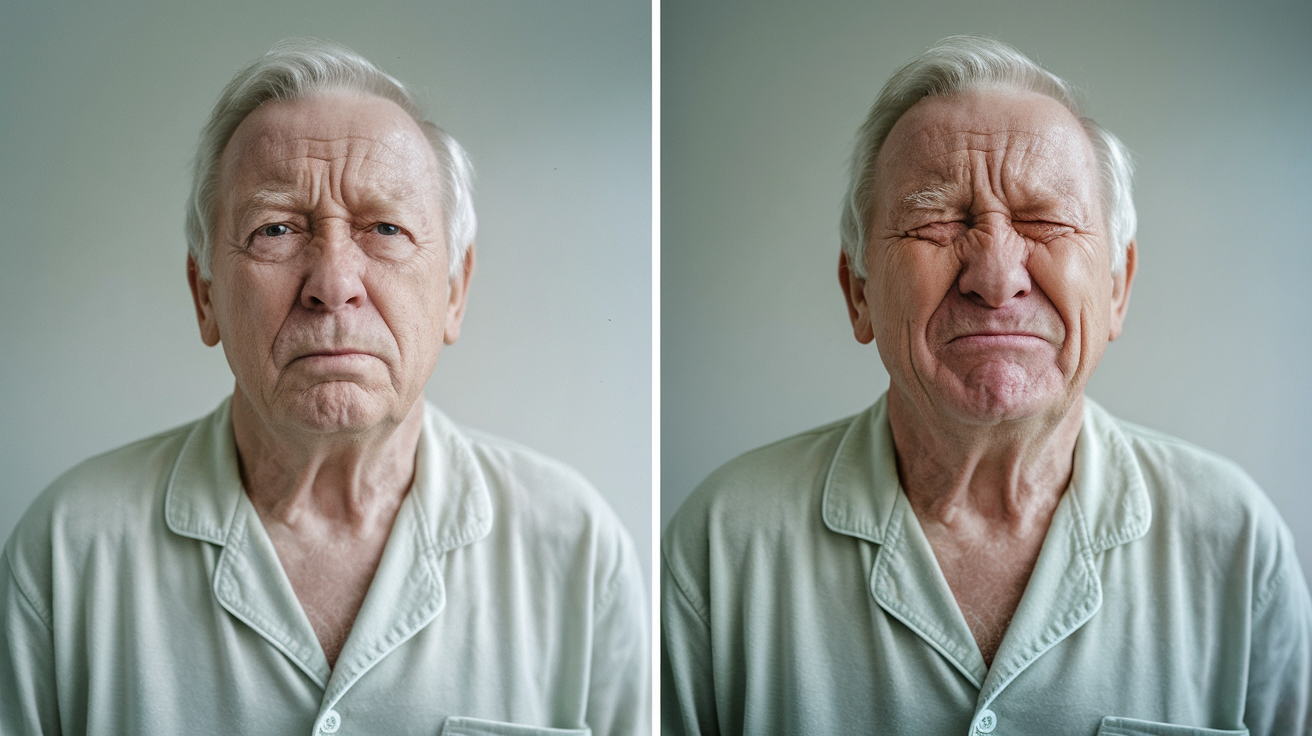}
    \includegraphics[width=.24\textwidth]
    {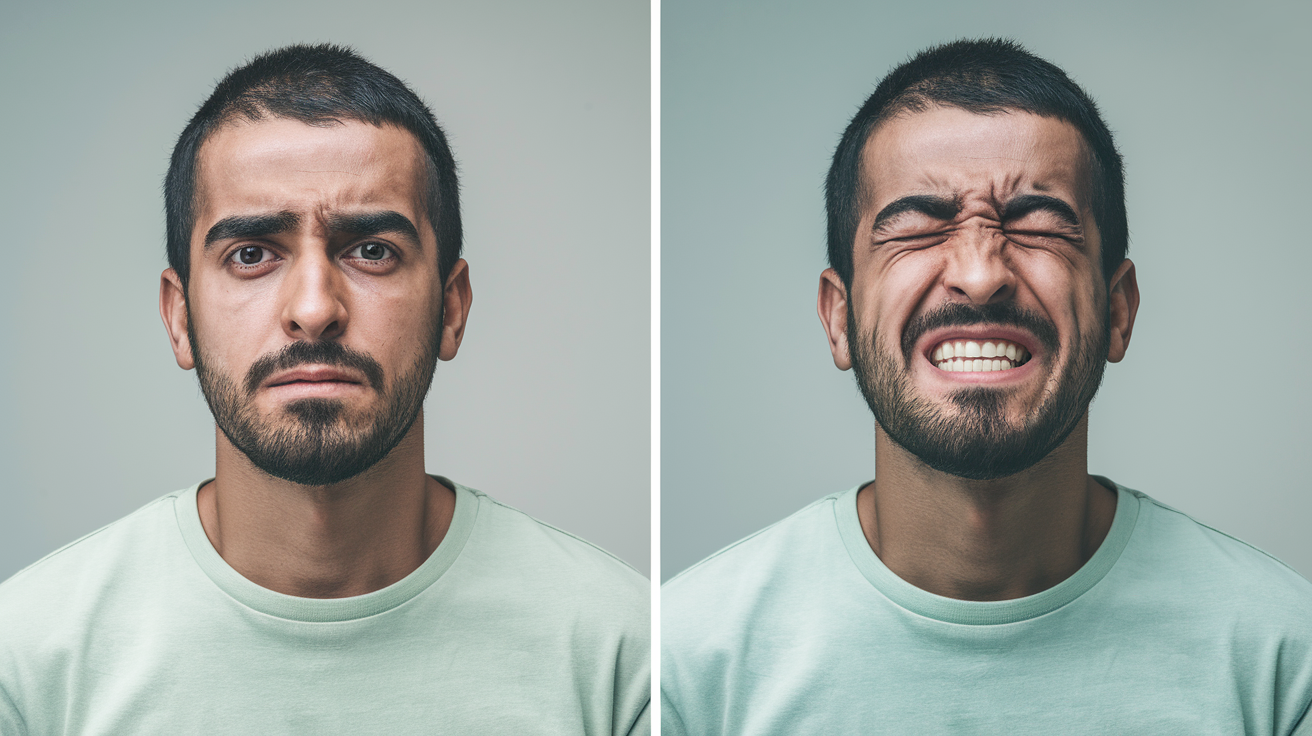}\\[1mm]
    \includegraphics[width=.24\textwidth]
    {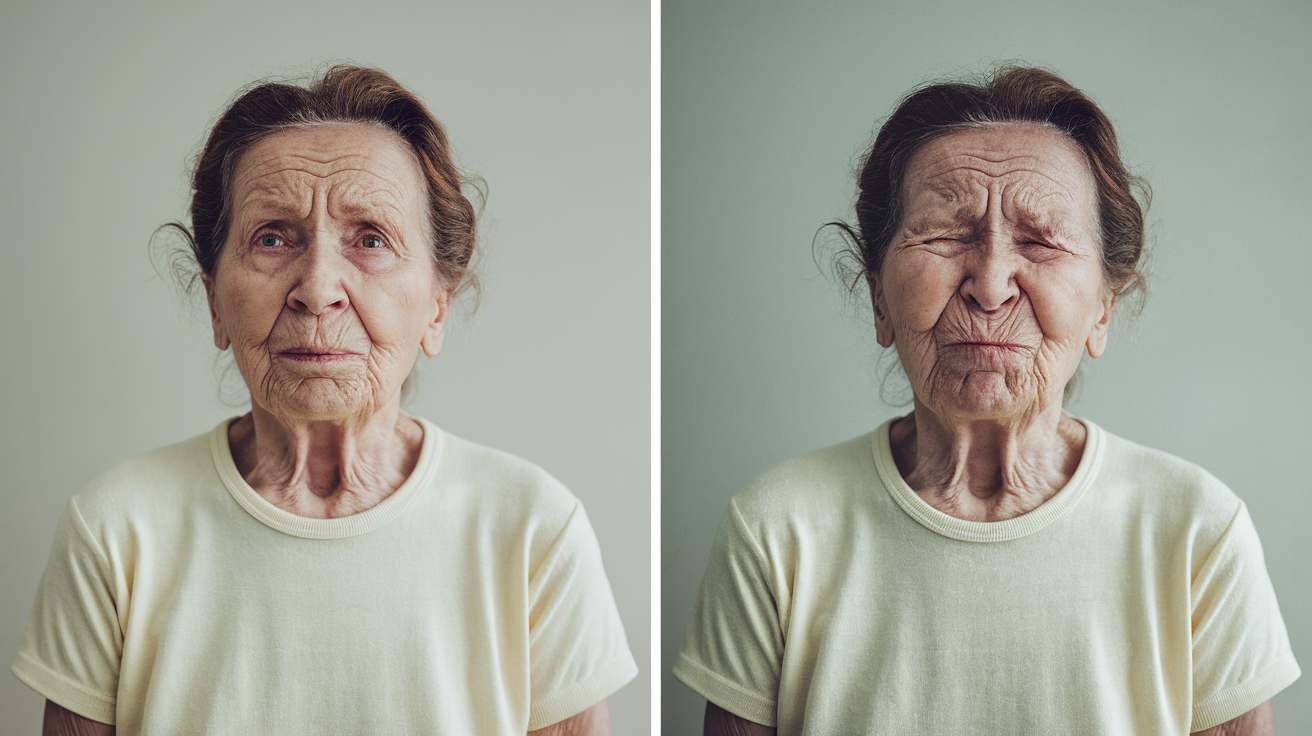}
    \includegraphics[width=.24\textwidth]
    {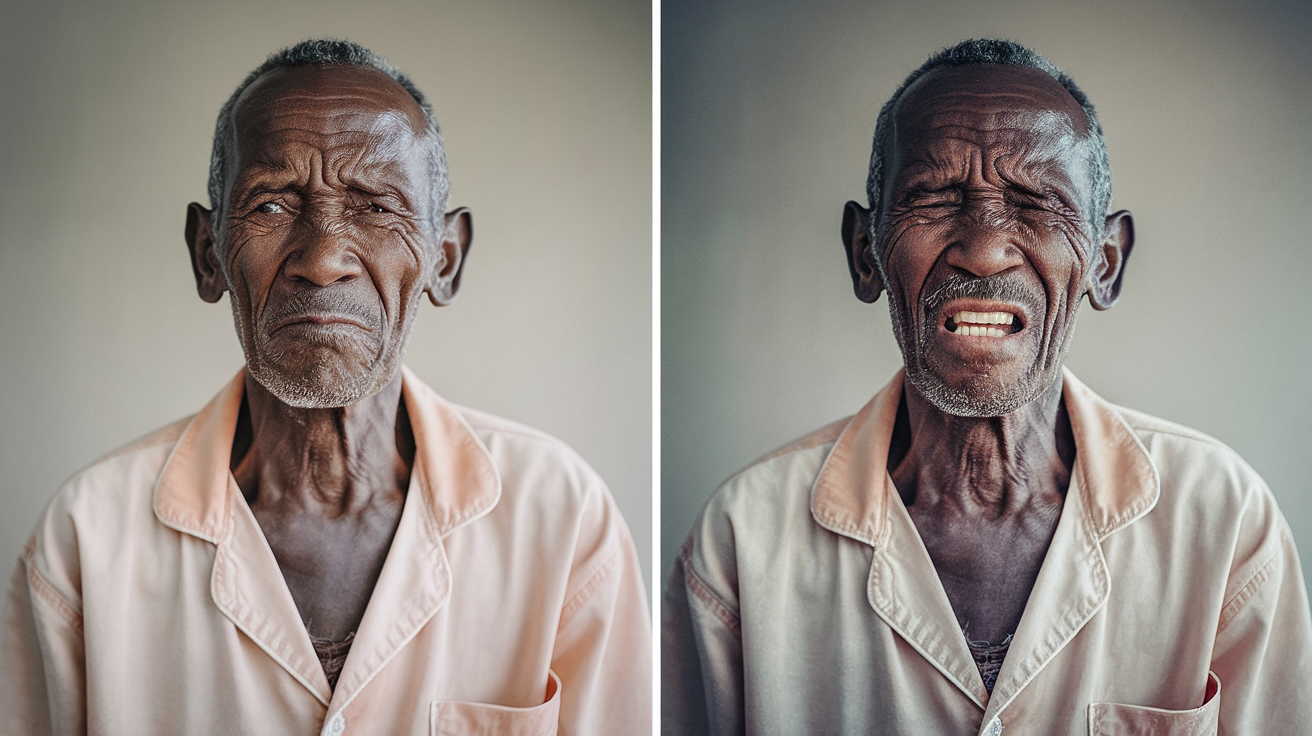}
    \includegraphics[width=.24\textwidth]
    {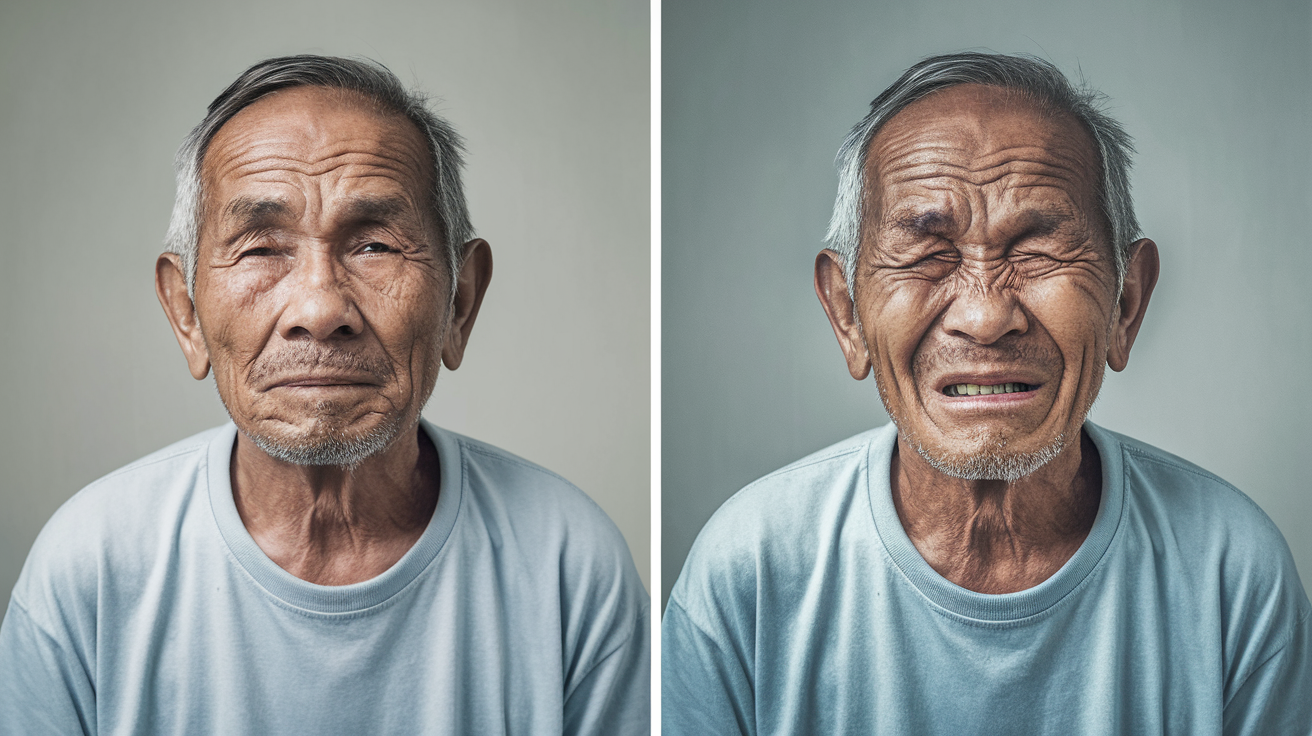}
    \includegraphics[width=.24\textwidth]
    {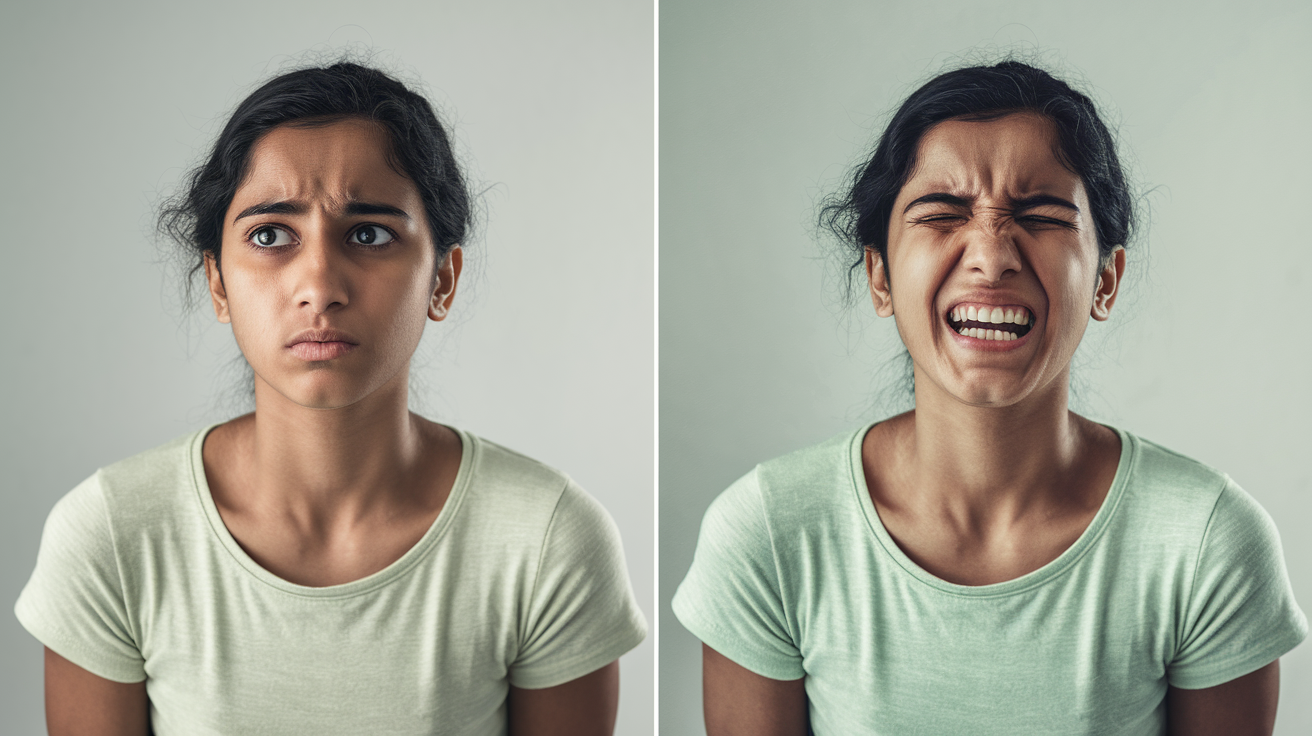}
    \caption{Sample image pairs from the \synpain dataset. {Left image in each pair is the neutral image}. Top row: Non-pain; bottom two rows: Pain; left three columns: Old; rightmost column: Young.}
    \label{fig:SampleSynPainImages}
\end{figure*}

\subsection{Expression Transfer}

One promising approach to augmenting pain detection training sets and increasing diversity is expression transfer~\cite{siarohin2019first,guo2024liveportrait,rochow2024fsrt,zhaox,PainControl}. Expression transfer can be used to superimpose real pain expressions (e.g., from public datasets) onto synthetic faces with varied attributes such as age, ethnicity, and gender. Recent work~\cite{PainControl} demonstrates that recent expression transfer methods can preserve synthetic target identities and successfully transfer pain expressions from a source image. However, when these synthetic images are used to augment pain detection training data, they do not improve downstream classifier performance~\cite{PainControl}. This may be because the transferred expressions are not truly novel, since the underlying pain expressions are still sourced from the same limited real datasets, or due to AU misalignment in the generated images. Regardless of the cause, these findings highlight a persistent need for datasets containing genuinely novel and demographically diverse pain and non-pain expressions. Entirely synthetic data generation, rather than superimposing real pain onto synthetic identities, is therefore necessary to overcome these limitations and advance the field.

\subsection{Face Data Synthesis}

{Recent advancements in generative AI have transitioned from the adversarial training of Generative Adversarial Networks (GANs) toward denoising diffusion probabilistic models (DDPMs), which offer superior mode coverage and high-fidelity facial synthesis through iterative refinement. To address the clinical requirement for precise affect control, state-of-the-art frameworks such as MagicFace~\cite{wang2024magicface} and FaceCrafter~\cite{mishima2025facecrafter} now utilize Action Units (AUs) and 3D blendshapes to achieve fine-grained, interpretable, and continuous expression editing while preserving subject identity. For longitudinal and subject-specific monitoring, identity-consistent foundation models like Arc2Face~\cite{paraperas2024arc2face} leverage discriminative embeddings to generate diverse images of specific individuals. These advancements in controllable image and video synthesis establish a robust foundation for augmenting training sets in data-scarce medical domains.}

\subsection{Contributions}

To address these issues, we present \synpain, a large, publicly available synthetic dataset of pain and non-pain facial expressions with annotated attributes for gender (male/female), ethnicity/race (5 groups), and age (young/old). We demonstrate that synthetic pain expressions exhibit clinically meaningful facial action unit patterns consistent with established pain assessment frameworks (PSPI) and establish \synpain's utility for examining algorithmic bias in existing pain detection models. Finally, we demonstrate that age-matched synthetic data augmentation improves pain detection performance on real clinical data.

\section{\synpain Dataset}

To support pairwise pain detection models~\cite{rezaei2020unobtrusive}, \synpain contains synthetic image pairs: one neutral expression and one expressive (pain or non-pain) image per identity. After qualitatively evaluating multiple generative AI tools, we selected Ideogram 2.0 (Ideogram, Inc., Toronto, Canada) for its superior facial detail and realistic expression synthesis. Using its paid API, we programmatically generated synthetic identities, each with paired neutral/expressive portrait images. {{For parameters, we specified an aspect ratio of 16:9, set the magic prompt option to ``OFF'', and provided a negative prompt of ``hat, glasses, headwear, scarf, hands, hand, profile, turban''. All remaining parameters were set to their default values. We did not generate the synthetic identities in a way to have three portrait images (i.e., one for each of neutral, pain, and non-pain) because when we prompted Ideogram 2.0 to do so, it returned a significant percent of images not following the given prompt or exhibiting strange artifacts.}

The dataset is annotated with the following attributes: Age: Young ({any age between 20 \& 35}) and Old ({any age {between 75 \& 93}}), Ethnicity/Race: White, Black, South Asian, East Asian, Middle Eastern, Gender: Male, Female, and Expression: Pain or Non-pain (e.g., talking, laughing). Each identity includes two aligned images (neutral/expressive), totaling 10,710 images (5,355 pairs). Roughly balanced number of images were generated for each attribute; but final numbers vary slightly after performing visual inspection to exclude the small number of images violating prompts (e.g., profile views) or those with artifacts or unrealistic expressions. ~\cref{tab:demographics} summarizes attribute distributions. ~\cref{fig:SampleSynPainImages} illustrates sample pairs from the dataset.

\begin{table}[]
\caption{Distribution of attributes in the \synpain dataset. Each cell contains the number of image pairs for the corresponding combination.}
\label{tab:demographics}
\centering
\begin{tabular}{lrrrrrr}
\toprule
                            & \textbf{Man} & \textbf{Woman} & \textbf{Pain} & \textbf{Not Pain} & \textbf{Total} \\ \midrule
\textbf{Young}              &  1220  &  1240  &  1146  &   1314    & 2460  \\
\textbf{Old}                &  1383  &  1512  &  1590  &   1305    & 2895  \\\midrule
\textbf{Man}                &  2603  &   0    &  1318  &   1285    & 2603  \\
\textbf{Woman}              &   0    &  2752  &  1418  &   1334    & 2752  \\\midrule
\textbf{Pain}               &  1318  &  1418  &  2736  &     0     & 2736  \\
\textbf{Not pain}           &  1285  &  1334  &   0    &   2619    & 2619  \\\midrule
\textbf{Black}              &  492   &  554   &  559   &    487    & 1046  \\
\textbf{White}              &  512   &  520   &  532   &    500    & 1032  \\
\textbf{Middle Eastern}     &  573   &  606   &  609   &    570    & 1179  \\
\textbf{South Asian}        &  507   &  541   &  519   &    529    & 1048  \\
\textbf{East Asian}         &  519   &  531   &  517   &    533    & 1050  \\\midrule
\textbf{Total}              &  2603  &  2752  &  2736  &   2619    & 5355  \\
\bottomrule
\end{tabular}
\end{table}

To encourage the model to generate variety, prompts varied clothing (e.g., ``She is wearing a [color] [garment]''), {weight (e.g., ``[He is thin/overweight]''),} and facial hair (e.g., ``[He has a beard]''). Attributes prompts also varied for age (e.g., ``She is [20-35] years old''), ethnicity/race (e.g., White, Caucasian, Greek, etc.), and expression. {The words in the prompt that varied ethnicity/race were primarily nationality labels (e.g., ``Spanish'' \& ``Dutch'' for ``White'', ``Nigerian'' \& ``Gabonese'' for ``Black'', etc.).} For pain expressions, this meant prompts ranging from simple descriptions (e.g., ``in pain'' or ``showing facial expressions of pain'') to a combination of descriptions relating to the FACS pain-related AUs used to derive the PSPI score or PACSLAC-II (e.g. ``lowered brow'', ``raised cheeks'', etc. from PSPI or ``groaning'', etc. from PACSLAC-II). {For non-pain expressions, prompts varied action (e.g., talking, laughing, smiling) and eye closure (e.g., open, closed, half-closed).}

Using RunwayML Gen-4 Alpha (Runway AI, Inc., New York, USA), we generated 5-second, 24~fps videos transitioning from neutral to expressive faces for 40 identities, representing one combination from each ethnicity/race, gender, expression type, and age group. \cref{fig:SampleVideoFrames} illustrates sample frames from one such generated video.

\begin{figure*}[t]
    \centering
    \begin{minipage}[t]{.07\textwidth}
        \centering
        \includegraphics[width=\linewidth]{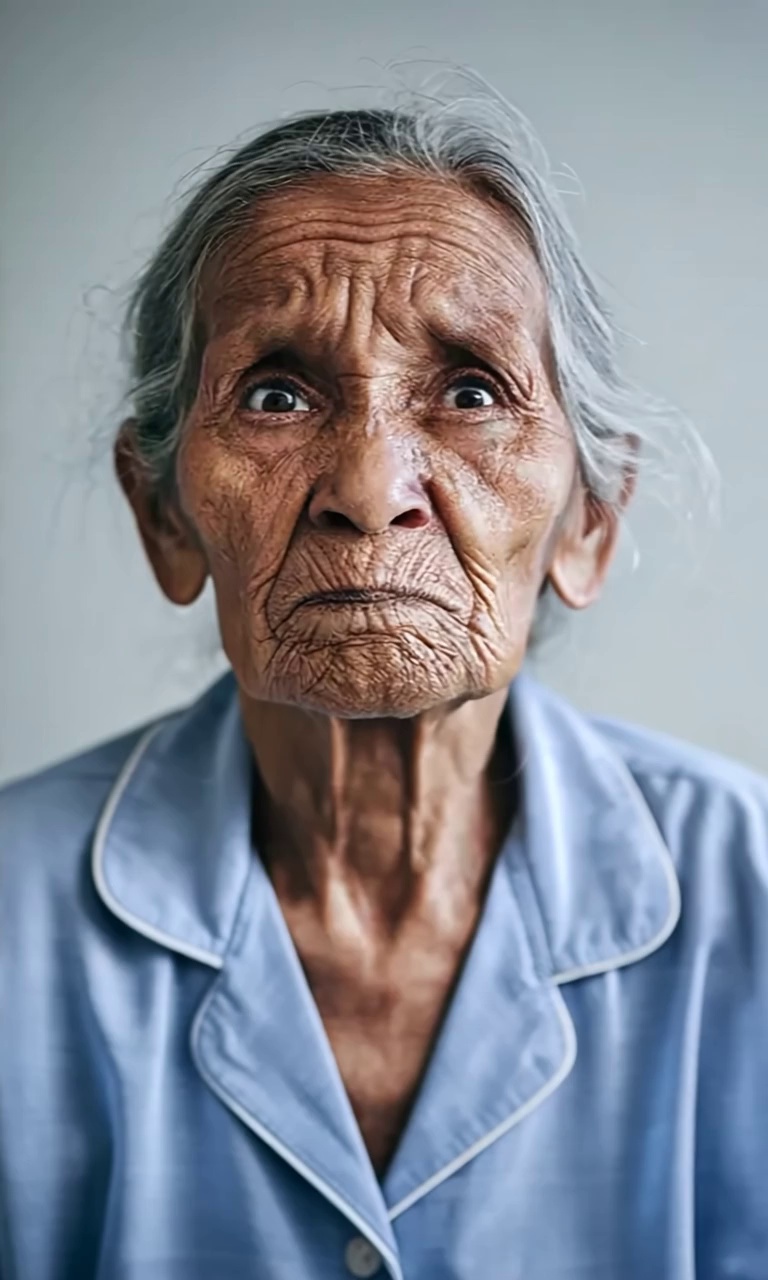}\\
        {\footnotesize 0.8}
    \end{minipage}
    \begin{minipage}[t]{.07\textwidth}
        \centering
        \includegraphics[width=\linewidth]{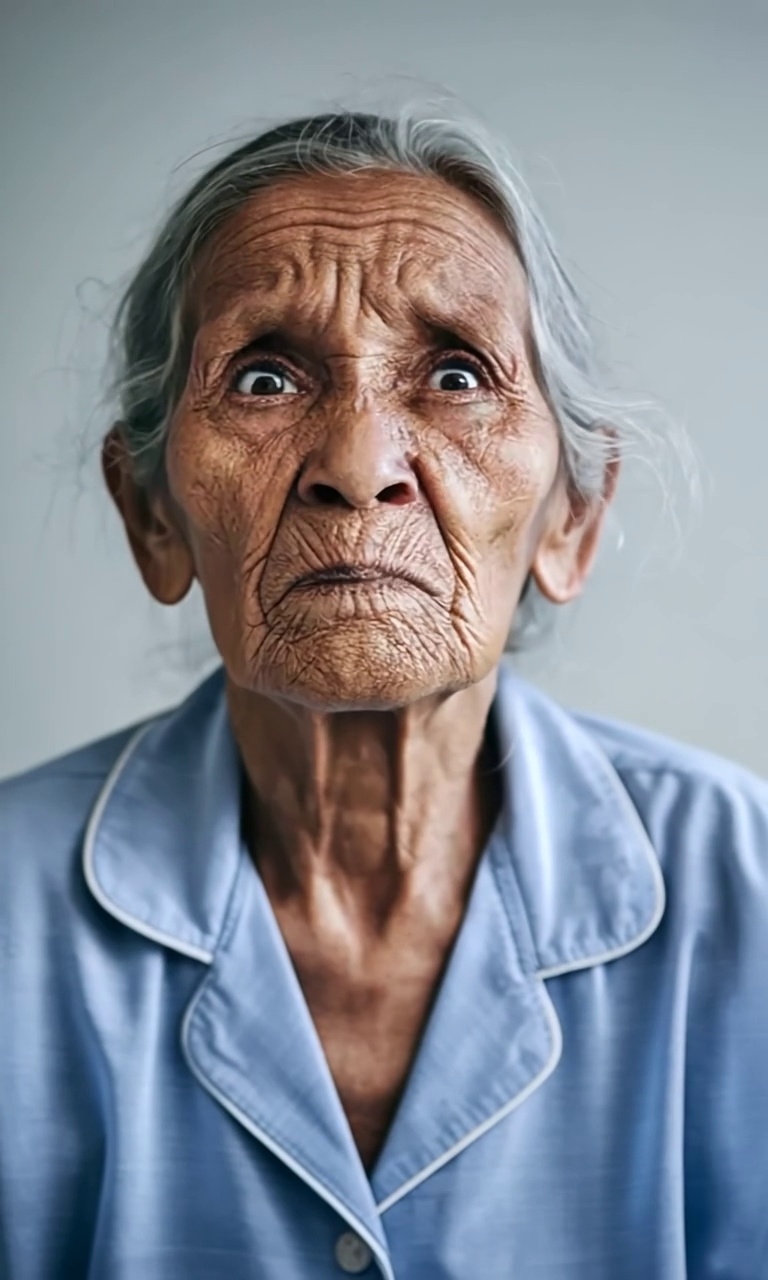}\\
        {\footnotesize 0.6}
    \end{minipage}
    \begin{minipage}[t]{.07\textwidth}
        \centering
        \includegraphics[width=\linewidth]{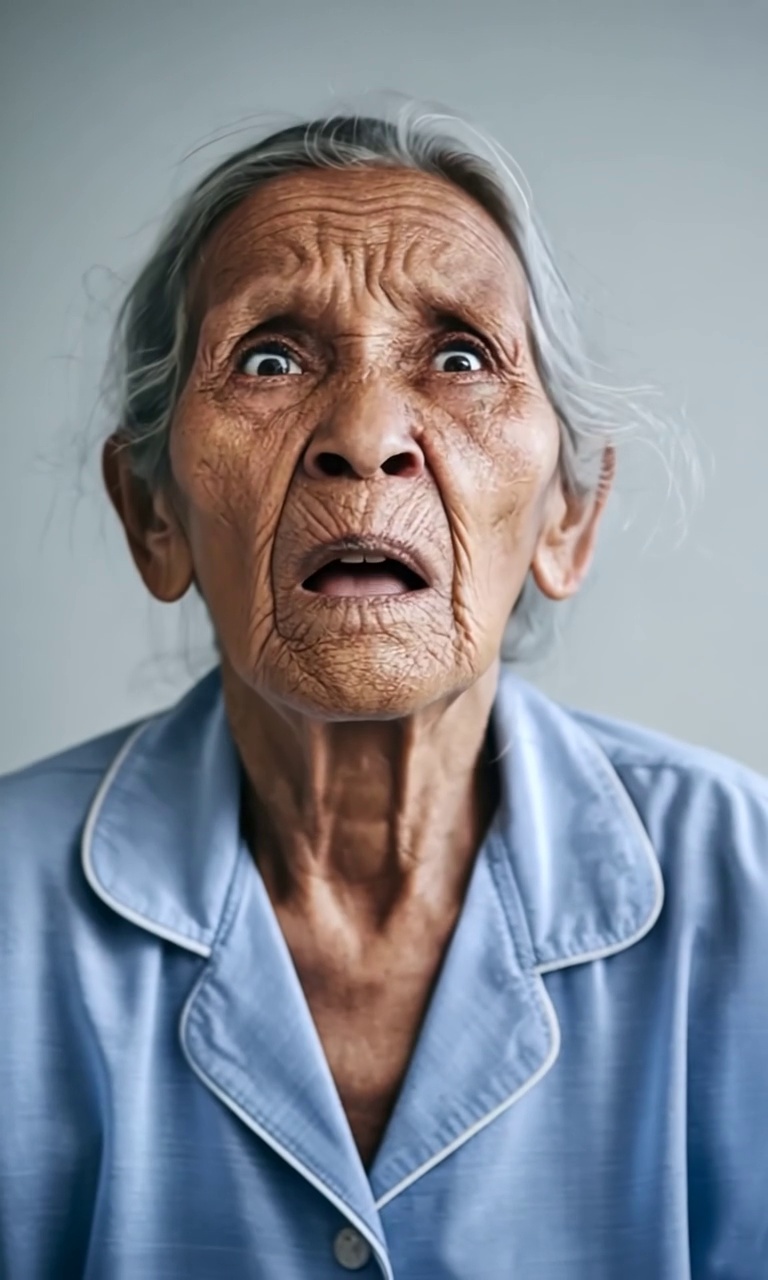}\\
        {\footnotesize 1.4}
    \end{minipage}
    \begin{minipage}[t]{.07\textwidth}
        \centering
        \includegraphics[width=\linewidth]{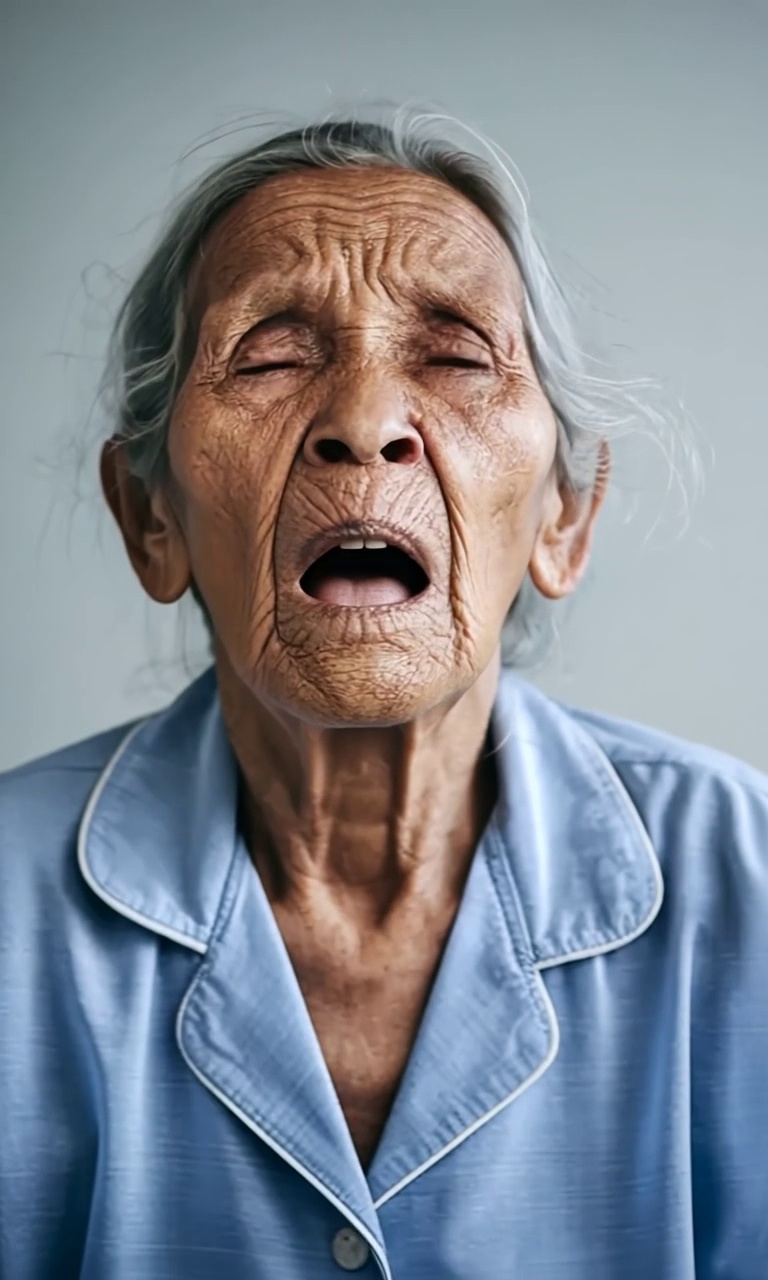}\\
        {\footnotesize 6.6}
    \end{minipage}
    \begin{minipage}[t]{.07\textwidth}
        \centering
        \includegraphics[width=\linewidth]{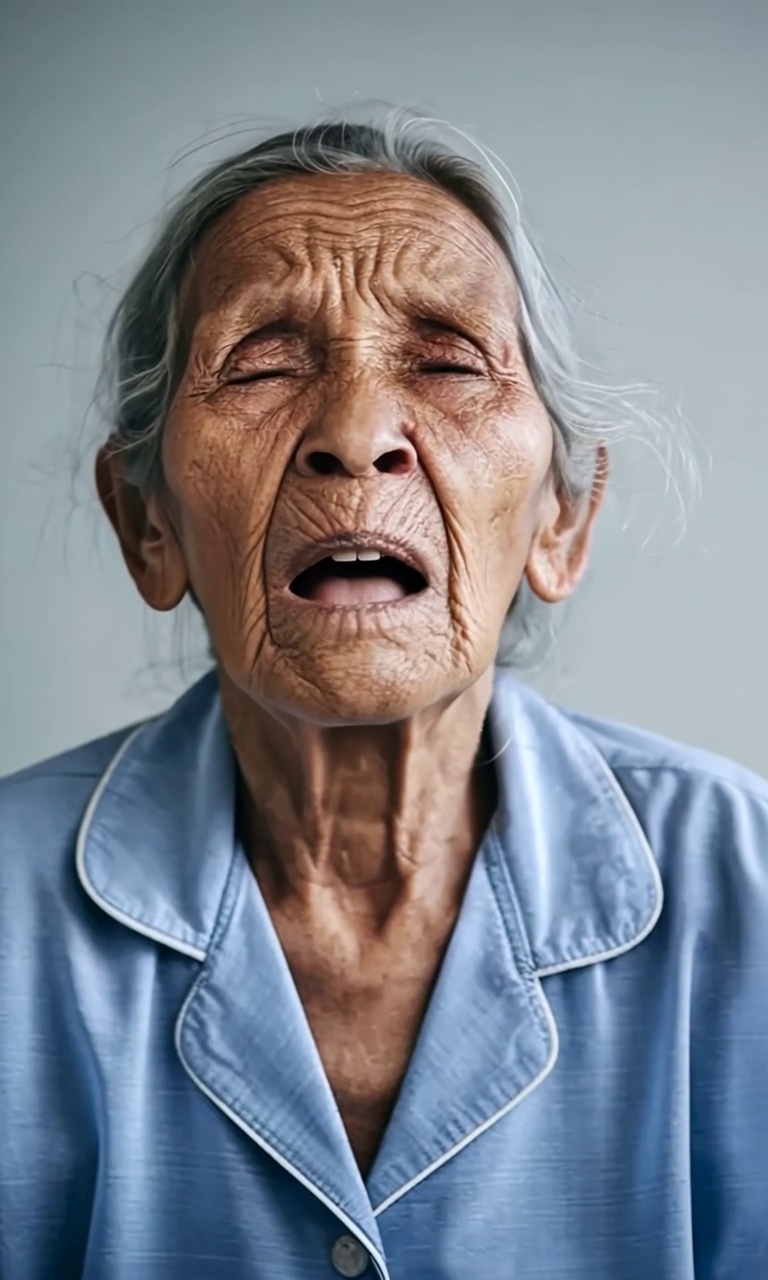}\\
        {\footnotesize 11.2}
    \end{minipage}
    \begin{minipage}[t]{.07\textwidth}
        \centering
        \includegraphics[width=\linewidth]{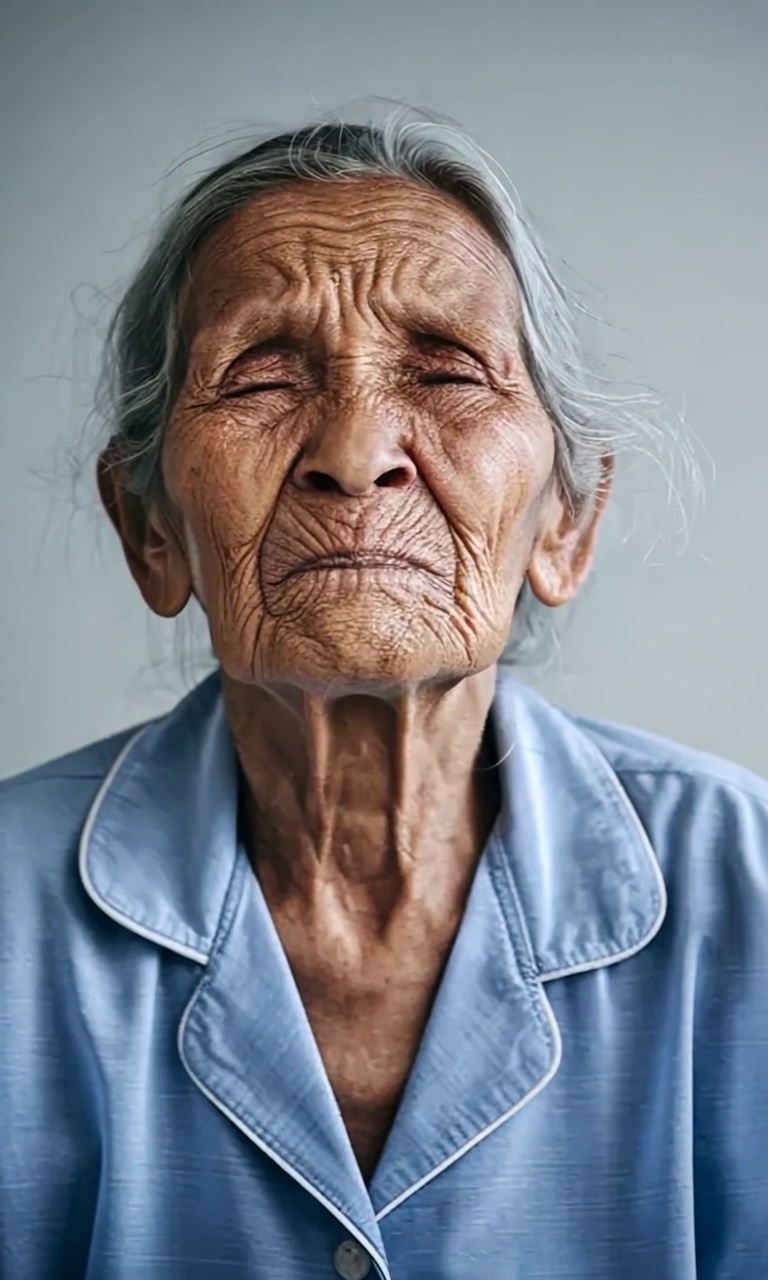}\\
        {\footnotesize 4.3}
    \end{minipage}
    \begin{minipage}[t]{.07\textwidth}
        \centering
        \includegraphics[width=\linewidth]{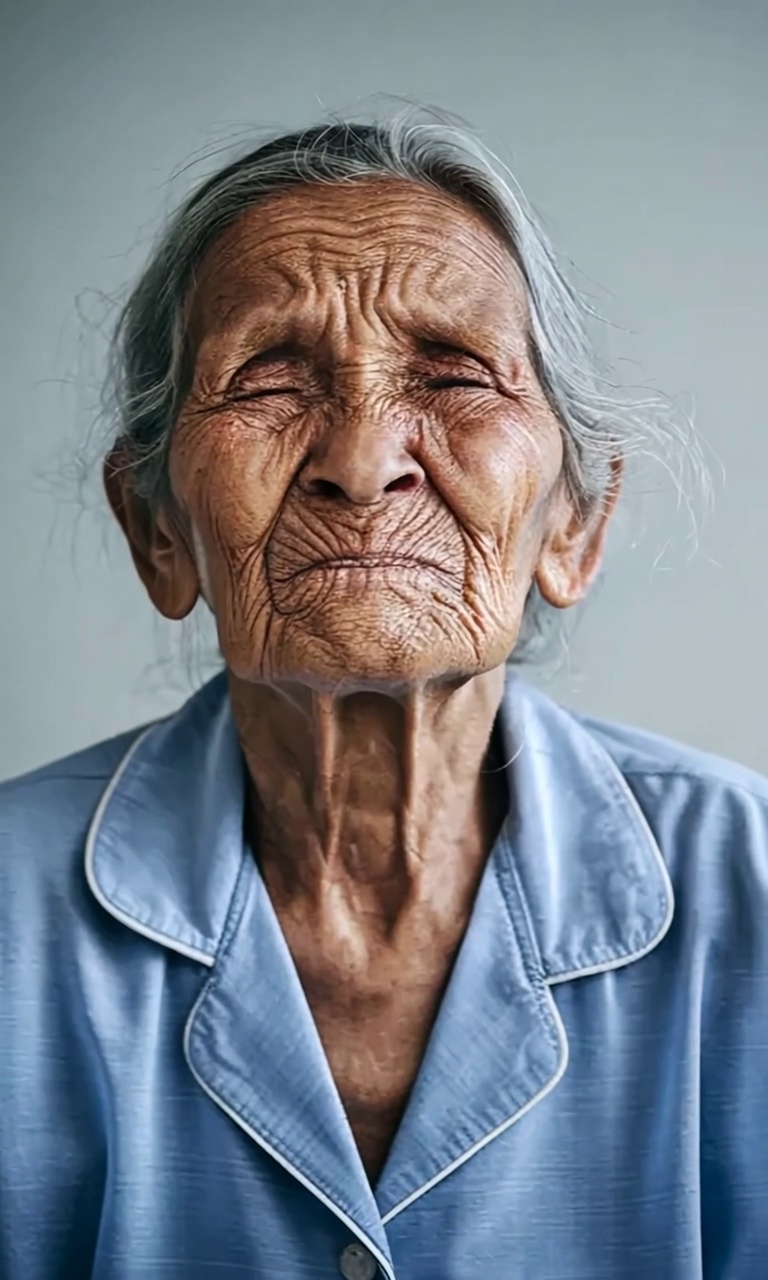}\\
        {\footnotesize 5.5}
    \end{minipage}
    \begin{minipage}[t]{.07\textwidth}
        \centering
        \includegraphics[width=\linewidth]{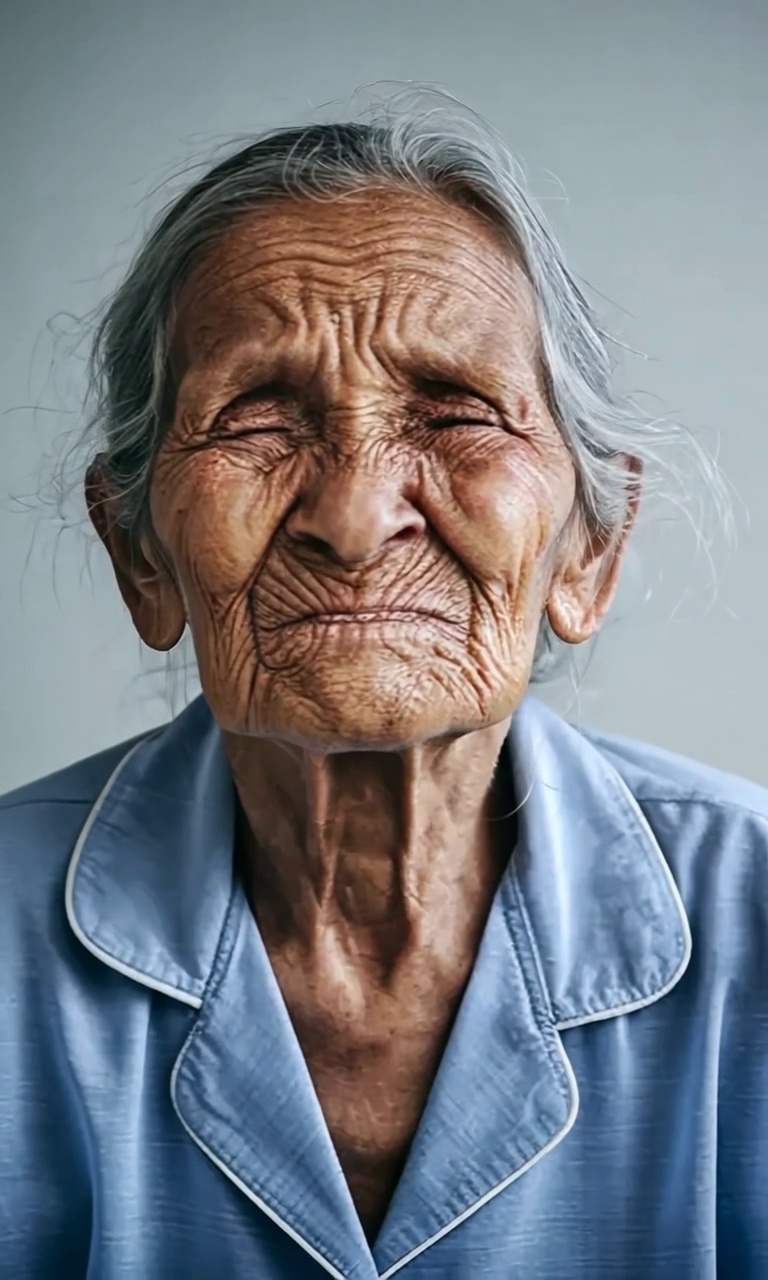}\\
        {\footnotesize 6.0}
    \end{minipage}
    \begin{minipage}[t]{.07\textwidth}
        \centering
        \includegraphics[width=\linewidth]{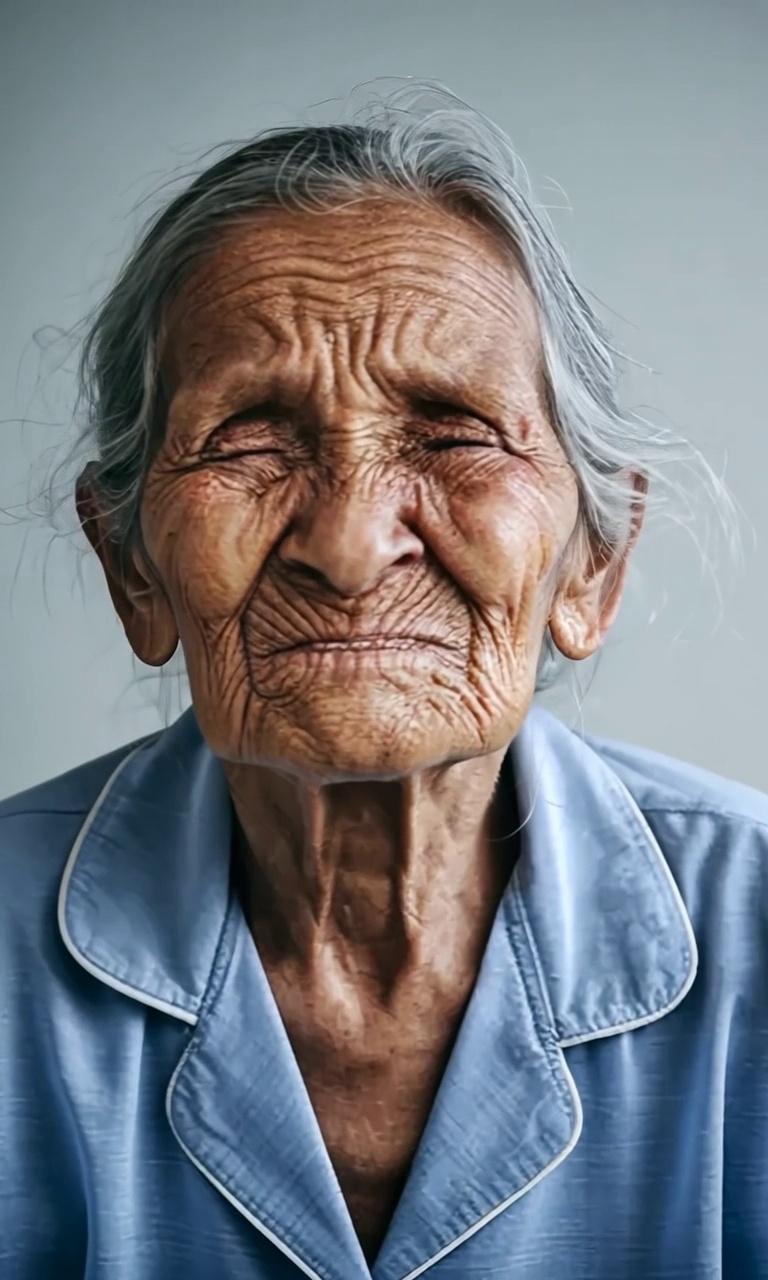}\\
        {\footnotesize 7.7}
    \end{minipage}
    \begin{minipage}[t]{.07\textwidth}
        \centering
        \includegraphics[width=\linewidth]{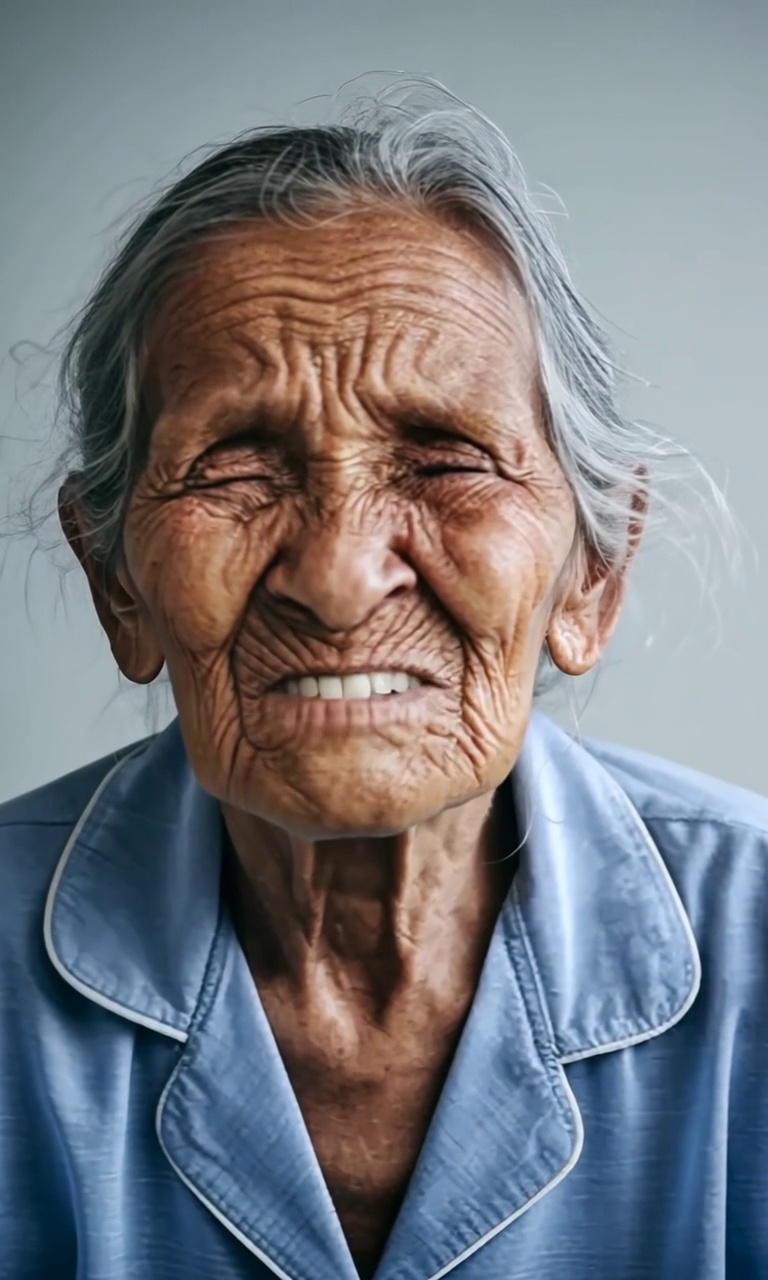}\\
        {\footnotesize 7.7}
    \end{minipage}
    \begin{minipage}[t]{.07\textwidth}
        \centering
        \includegraphics[width=\linewidth]{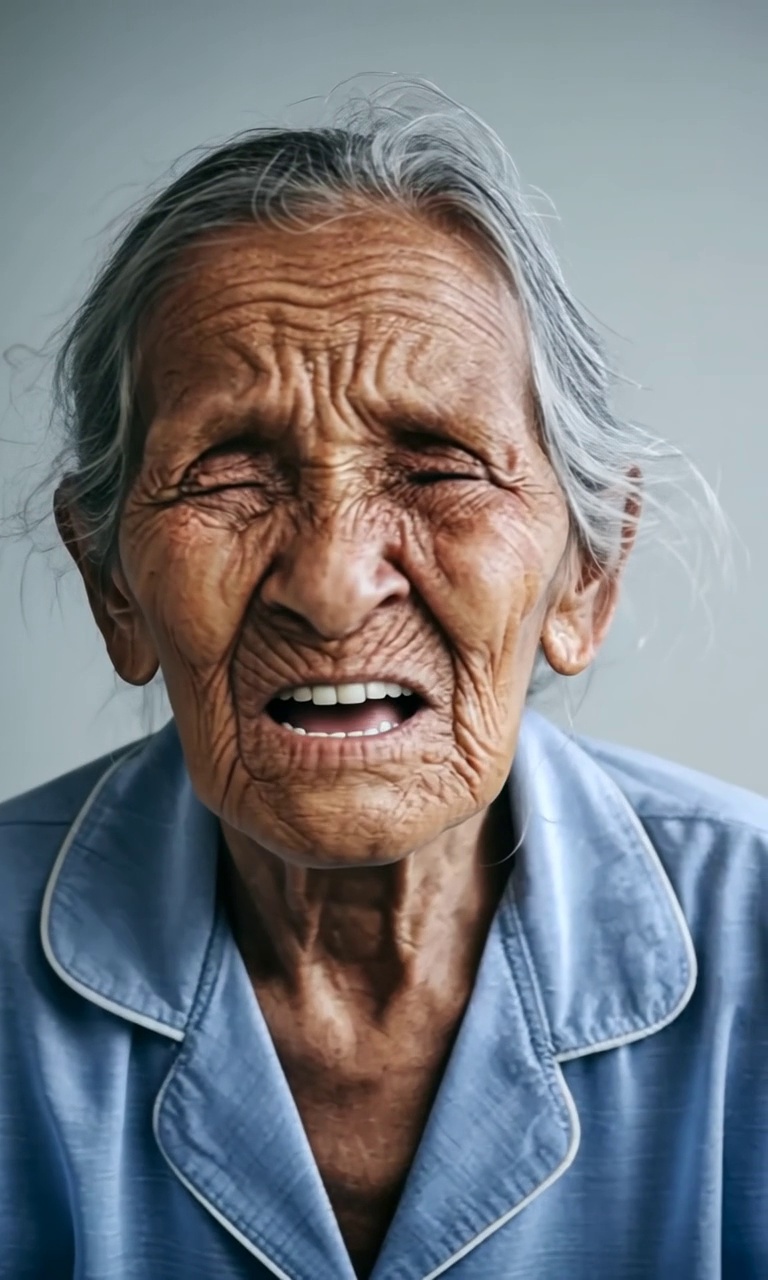}\\
        {\footnotesize 9.9}
    \end{minipage}
    \begin{minipage}[t]{.07\textwidth}
        \centering
        \includegraphics[width=\linewidth]{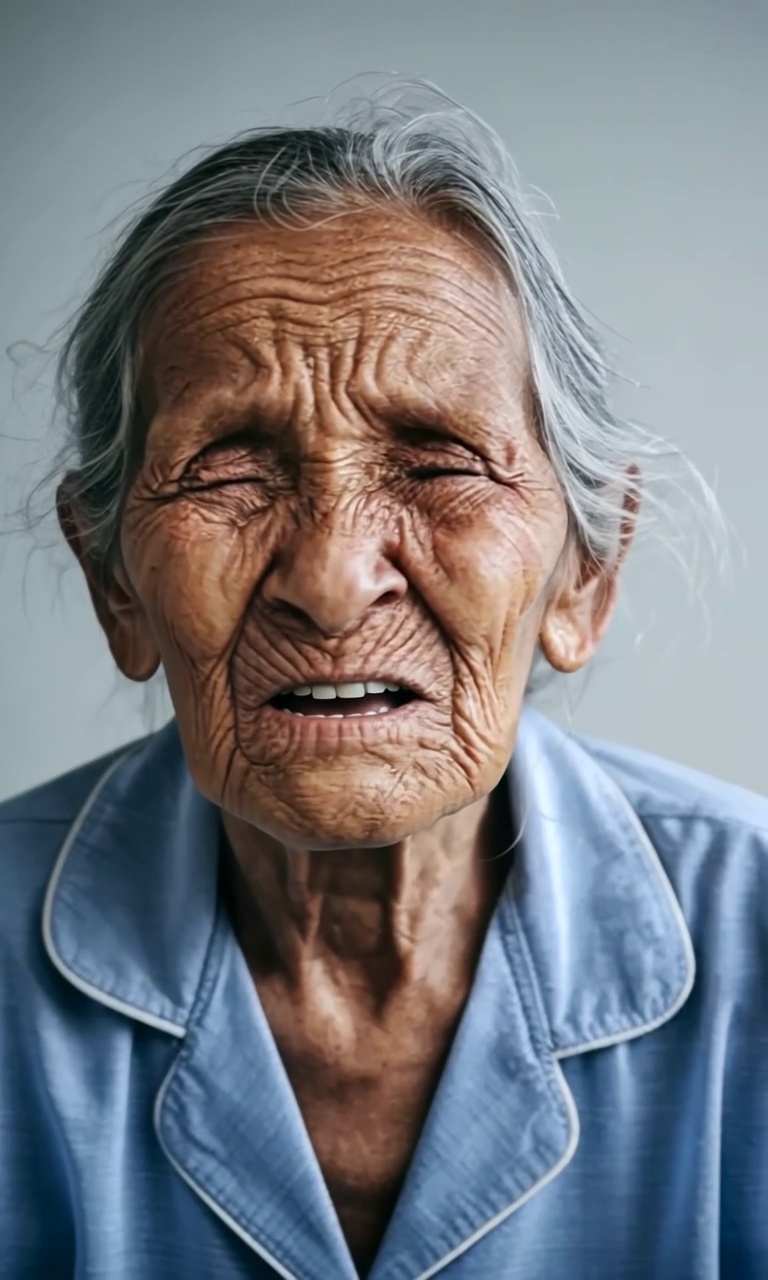}\\
        {\footnotesize 9.6}
    \end{minipage}
    \caption{Sample frames from a 5-second video showing the progression from a neutral expression to a facial expression of pain. The number below shows the estimated pain level (PSPI, in the [0,16] range) for each frame, as detected by the pretrained PwCT model~\cite{rezaei2020unobtrusive}.}
    \label{fig:SampleVideoFrames}
\end{figure*}

\subsection{Quality of Generated Images}

We evaluated the quality of \synpain images using DSL-FIQA~\cite{chen2024dsl}, a state-of-the-art method for assessing facial image quality via dual-set degradation learning and landmark-guided transformers. DSL-FIQA provides quality scores in the range [0, 1], where higher scores indicate better image quality. Following standard practice, we interpret these scores according to the ACR (Absolute Category Rating) scale: Bad (0-0.2), Poor (0.2-0.4), Fair (0.4-0.6), Good (0.6-0.8), and Excellent (0.8-1.0). 

\cref{tab:facequality} shows the average face quality scores for all \synpain images (10,710 images total from 5,355 pairs). The average quality score for neutral images was 0.865, while expressive images achieved an average score of 0.871, both falling within the excellent quality range of the ACR scale. 

To corroborate these findings, we also evaluated image quality using Py-Feat~\cite{cheong2023py}, which provides face detection confidence scores in the 0-1 range. The results demonstrate consistently high quality across all images: neutral images achieved a mean score of 0.999 (SD = 0.002), while expressive images achieved a mean score of 0.998 (SD = 0.004). The high median values ($>$0.999 for both conditions) and narrow interquartile ranges ($<$0.002) confirm that the vast majority of generated images meet high quality standards for facial analysis applications.

To verify our visual quality control process, we further numerically assessed head pose alignment using Py-Feat~\cite{cheong2023py} to estimate roll, pitch, and yaw angles across all 10,710 images in \synpain. The analysis confirms excellent frontal alignment: only 0.1\% of images exhibited excessive pitch rotation ($\left| pitch \right| > 20\degree$) and 1.2\% showed excessive yaw rotation ($\left| yaw \right|>20\degree$). 
These results validate our visual inspection process and demonstrate that \synpain predominantly consists of well-aligned frontal face images suitable for standard face analysis pipelines.

\begin{table}[]
\caption{DSL-FIQA face quality scores of \synpain images. Each cell presents the average quality score (0-1 scale) for all images in the corresponding demographic and expression combination.}
\label{tab:facequality}
\centering
\begin{tabular}{lccccc}
\toprule
                            & \textbf{Man} & \textbf{Woman} & \textbf{Pain} & \textbf{Not Pain} & \textbf{All} \\ \midrule
\textbf{All}                &  0.875  &  0.860  &  0.875  &   0.859    & 0.868  \\\midrule
\textbf{Young}              &  0.875  &  0.854  &  0.872  &   0.858    & 0.865  \\
\textbf{Old}                &  0.875  &  0.865  &  0.878  &   0.860    & 0.870  \\\midrule
\textbf{Man}                &  0.875  &   - &  0.881  &   0.869    & 0.875  \\
\textbf{Woman}              &  -  &  0.860  &  0.870  &   0.850    & 0.860  \\\midrule
\textbf{Pain}               &  0.881  &  0.870  &  0.875  &   -    & 0.875  \\
\textbf{Not pain}           &  0.869  &  0.850  &  -  &   0.859    & 0.859  \\\midrule
\textbf{Black}              &  0.874  &  0.865  &  0.879  &   0.858    & 0.869  \\
\textbf{White}              &  0.872  &  0.866  &  0.876  &   0.861    & 0.869  \\
\textbf{Middle Eastern}     &  0.879  &  0.860  &  0.878  &   0.860    & 0.869  \\
\textbf{South Asian}        &  0.876  &  0.855  &  0.871  &   0.860    & 0.866  \\
\textbf{East Asian}         &  0.874  &  0.856  &  0.872  &   0.857    & 0.865  \\
\bottomrule
\end{tabular}
\end{table}

\subsection{Perceived Demographics of the Dataset}

{To validate the demographic and expression labels in our dataset, we conducted a user study with two independent annotators who classified perceived gender, age, race/ethnicity, and facial expression for all samples. As shown in Table~\ref{tab:userstudy}, both annotators achieved perfect agreement on gender and age classifications, with 100\% accuracy and a Fleiss' Kappa of 1.00, indicating complete inter-rater reliability for these categories. This perfect agreement suggests that the gender and age labels in our dataset are unambiguous and consistently perceivable across different observers.}

{For the more subjective categories of expression and race/ethnicity, we observed slightly lower but still strong agreement. Expression classification achieved 96.0\% and 94.0\% accuracy for the two annotators respectively, with a Fleiss' Kappa of 0.94, indicating excellent inter-rater reliability. Race/ethnicity classification showed 93.0\% and 90.5\% accuracy with a Fleiss' Kappa of 0.92, also demonstrating excellent agreement. The marginally lower agreement for race/ethnicity likely reflects the inherent complexity and subjectivity of racial categorization based on visual appearance alone, while the high kappa values overall confirm that our dataset labels are reliable and consistently interpretable across independent human raters.}

\begin{table}[]
\caption{{User study results for demographic and expression classification. Each cell presents the classification accuracy for the corresponding category.}}
\label{tab:userstudy}
\centering
{
\begin{tabular}{lccc}
\toprule
                            & \textbf{User 1} & \textbf{User 2}  & \textbf{Fleiss' Kappa}\\ \midrule
\textbf{Gender}                &   100.0\%&   100.0\% &1.00\\
\textbf{Age}                &   100.0\%&   100.0\% &1.00\\
\textbf{Expression}                &   96.0\%&   94.0\% & 0.94 \\
\textbf{Race/Ethnicity}                &   93.0\%&   90.5\% & 0.92 \\
\bottomrule
\end{tabular}
}
\end{table}

\subsection{Validity of the Pain Expressions}

To demonstrate the validity of the pain expressions in \synpain, we used off-the-shelf AU detectors to calculate PSPI scores and compared pain versus non-pain images. While direct pain detection (training end-to-end pain detection models) achieves superior performance compared to AU-based approaches~\cite{rezaei2020unobtrusive}, this AU-based validation provides a straightforward verification that our synthesized pain expressions exhibit the expected facial action unit patterns.

We used FaceReader Version 9.1 (Noldus Information Technology, Netherlands), a commercial AU detection system, to analyze all neutral and expression images in \synpain. \cref{tab:facereaderfail} shows the failure rates of FaceReader's AU detection across different ethnicities/races and pain conditions. {``Failure'' here refers to FaceReader's failure to detect a face, which, as a result, led to no AUs being detected.} The results reveal that performance is significantly influenced by {age, gender, skin tone, and the presence of pain expressions}. 
{ A chi-square test of independence revealed significant differences in failure rates across ethnic groups ($\chi^2(4) = 417.04$, $p < 0.001$, Cramér's $V = 0.20$). The results reveal that performance is significantly influenced by skin tone, with FaceReader failing substantially more on Black faces (11.4\% failure rate) compared to White (1.0\%), East Asian (0.8\%), and Middle Eastern (2.2\%) faces. Post-hoc pairwise comparisons with Bonferroni correction ($\alpha$ = 0.05) confirmed that failure rates differed significantly between all ethnic pairs (all $p < 0.05$) except White versus East Asian faces. Compared to White faces as reference, Black faces showed 13.1 times higher odds of failure (OR = 13.12, 95\% CI [8.28, 20.80], $p < 0.001$), South Asian faces showed 6.5 times higher odds (OR = 6.48, 95\% CI [4.03, 10.43], $p < 0.001$), and Middle Eastern faces showed 2.3 times higher odds (OR = 2.35, 95\% CI [1.40, 3.94], $p = 0.013$). East Asian faces showed no significant difference from White faces (OR = 0.83, 95\% CI [0.44, 1.60], $p = 1.00$).}
Additionally, the system fails considerably more on images containing pain expressions (12.4\%) compared to neutral (1.2\%) or non-pain expressions (1.9\%). The first observation (varying performance across ethnicities/races) aligns with previous findings~\cite{buolamwini2018gender} regarding algorithmic bias in facial analysis systems with respect to skin tone, while the latter finding (poor performance on pain expressions) is consistent with our understanding that pain expressions are underrepresented in public datasets, meaning commercial models have had limited exposure to such facial configurations. 
{ FaceReader also showed significantly higher failure rates for older adult faces 
(6.84\% vs 1.16\% for young adults; OR = 6.26, $p < 0.001$) and male faces (6.59\% vs 2.00\% for female faces; OR (Man vs. Woman) = 3.46, $p < 0.001$) (Table~\ref{tab:age_gender_failures}).}
{To examine whether the effects of ethnicity/race and gender were independent or interacted, we analyzed failure rates across all ethnicity/race-gender combinations (Table~\ref{tab:ethnicity_gender_interaction}). White and East Asian female faces showed extremely low failure rates (0.10\% and 0.19\% respectively) compared to their male counterparts (1.86\% and 1.45\%), though the small number of failures in these groups (1 and 2 failures, respectively) precludes precise estimation of the magnitude of these differences, resulting in very wide confidence intervals (e.g., OR = 19.64, 95\% CI [2.62, 147.00] for 
White faces). Failure rates varied significantly across ethnicity-gender combinations ($\chi^2(9) = 708.42$, $p < 0.00$, Cramér's $V = 0.26$). A logistic regression analysis confirmed a significant ethnicity $\times$ gender interaction ($\chi^2(4) = 33.57$, $p < 0.001$), indicating that gender effects on failure rates varied across ethnic groups.}
These results already demonstrate both the necessity and utility of \synpain in addressing these gaps and identifying systematic biases in existing facial analysis tools.

For the 95.8\% of \synpain images where FaceReader successfully detected PSPI-related AUs, we calculated PSPI scores and performed statistical comparisons using unpaired Mann-Whitney U tests between neutral, non-pain expression, and pain expression images. Results, shown in \cref{fig:AUdetection}, confirm that the mean estimated PSPI values were lowest for neutral faces (2.9), followed by non-pain expressions (4.3), and highest for pain expressions (6.7). All pairwise differences were statistically significant ($p<10^{-5}$).

\begin{figure}[t]
    \includegraphics[width=.9\columnwidth]{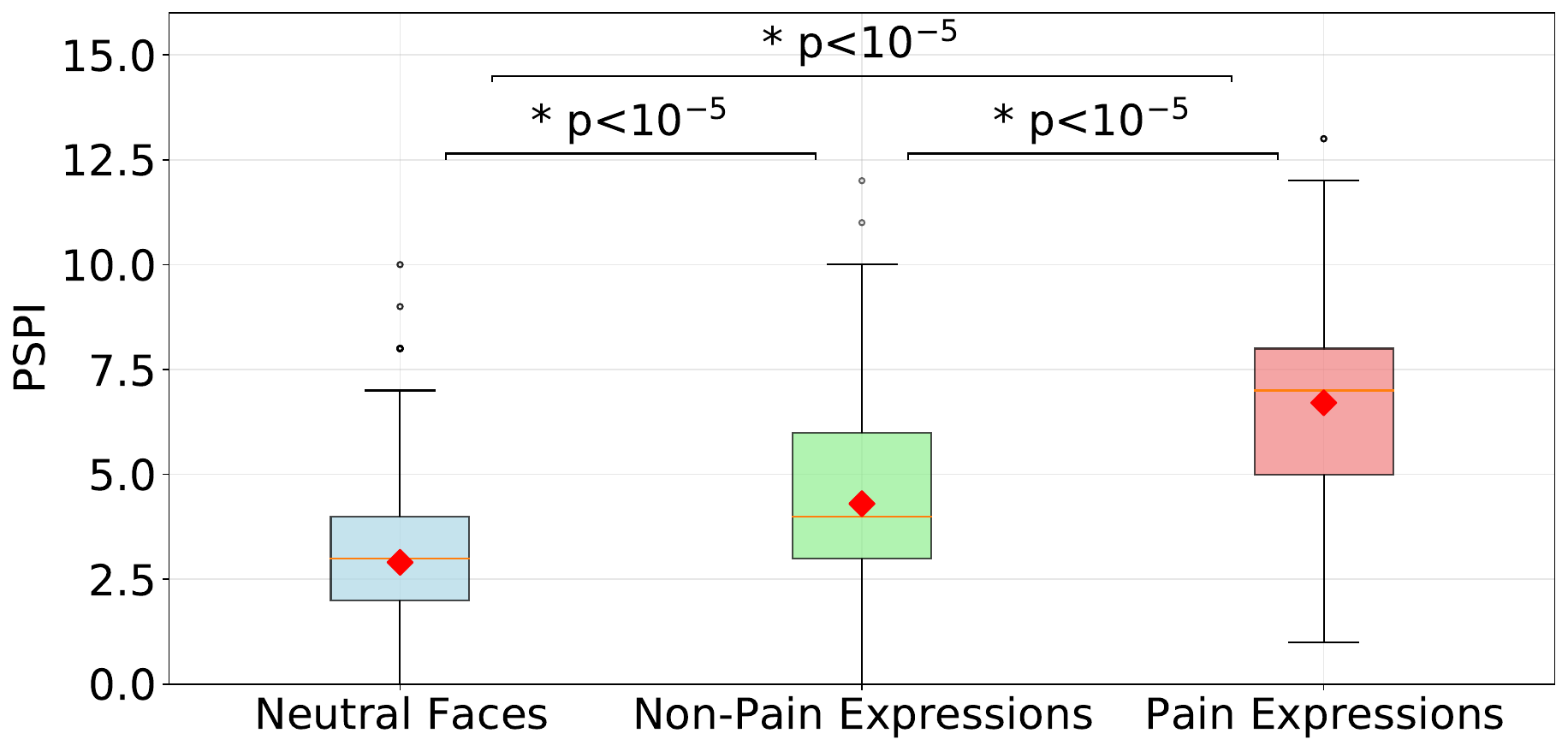}
    \caption{Distribution of PSPI scores calculated from FaceReader AU detections across expression conditions. Statistical significance was assessed using unpaired Mann-Whitney U tests.}  
    \label{fig:AUdetection}
\end{figure}

\begin{table}[]
\centering
\caption{Percentage of images for which FaceReader failed to perform AU detection by ethnicity/race and expression condition.}
\label{tab:facereaderfail}
\centering
\begin{tabular}{lrrrrr}
\toprule
               & \textbf{Neutral} & \textbf{Not Pain} & \textbf{Pain} & \textbf{All}  \\\midrule
\textbf{Black}          & 3.7     & 6.6      & 29.9 & 11.4 \\
\textbf{White}    & 0.0     & 0.2      & 3.6  & 1.0  \\
\textbf{South Asian} & 2.3     & 2.3      & 17.1 & 6.0  \\
\textbf{Middle Eastern}          & 0.1     & 0.7      & 7.9  & 2.2  \\
\textbf{East Asian}     & 0.0     & 0.0      & 3.3  & 0.8  \\\midrule
\textbf{Total}          & 1.2     & 1.9      & 12.4 & 4.2 \\
\bottomrule
\end{tabular}
\end{table}

\begin{table}[h]
\centering
\caption{{FaceReader AU detection failure rates by age and gender.}}
\label{tab:age_gender_failures}
{
\begin{tabular}{lcc}
\toprule
                & \textbf{Failure Rate (\%)} & \textbf{OR (95\% CI)} \\   \midrule
\textbf{Young}  & 1.16 & 1.00 (Reference) \\
\textbf{Old}    & 6.84 & 6.26 (4.73, 8.29) \\\midrule
\textbf{Woman}  & 2.00 & 1.00 (Reference) \\
\textbf{Man}    & 6.59 & 3.46 (2.78, 4.30) \\
\bottomrule
\end{tabular}
}
\end{table}

\begin{table}[h]
\centering
\caption{{Ethnicity × Gender interaction in FaceReader failure rates.\\(ME: Middle Eastern, SA: South Asian, EA: East Asian)}}
\label{tab:ethnicity_gender_interaction}
{
\begin{tabular}{lcccc}
\toprule
                        & \textbf{Man \%} & \textbf{Woman \%} & \textbf{Diff} & \textbf{OR (95\% CI)} \\\midrule
\textbf{Black}          & 18.80 & 4.78 & 14.02 & 4.61 (3.35, 6.34) \\
\textbf{White}          & 1.86 & 0.10 & 1.76 & 19.64 (2.62, 147.00) \\
\textbf{ME} & 4.28 & 0.33 & 3.95 & 13.49 (4.85, 37.50) \\
\textbf{SA}    & 7.40 & 4.62 & 2.78 & 1.65 (1.14, 2.38) \\
\textbf{EA}     & 1.45 & 0.19 & 1.26 & 7.77 (1.77, 34.07) \\
\bottomrule
\end{tabular}
}
\end{table}

\subsection{Identities}
{We used Py-Feat~\cite{cheong2023py} to encode facial identities and assess identity consistency within \synpain. Our evaluation had two goals: first, to confirm that neutral and expressive images within each pair appear to come from the same person (identity preservation); and second, to verify that faces within a given gender-race/ethnicity cell are not more similar to each other than faces in other demographic cells (demographic diversity). To evaluate identity preservation}, we compared cosine similarities between: (1) neutral and expressive images of the same identity (matching pairs), and (2) neutral images from different identities (non-matching pairs).
The results demonstrate strong identity consistency in our synthetic dataset. Matching pairs achieved a median cosine similarity of 0.72, while non-matching pairs showed significantly lower similarity with a median of 0.19. Statistical analysis confirmed this difference is highly significant (Mann-Whitney U test, $p<0.001$) with a large effect size (Cohen's d = 2.45).
{To benchmark these results against photographic stimuli, we repeated the same analysis on 100 matching pairs (neutral and expressive faces from the same identity) and 100 non-matching pairs (neutral faces from different identities) drawn from the Chicago Face Database (CFD)~\cite{ma2105cfd} The CFD yielded median cosine similarities of 0.835 and 0.257 for matching and non-matching pairs, respectively (Cohen's d = 2.79). The effect size observed in SynPAIN (d = 2.45) is comparable to that obtained with real photographs, indicating that our synthetic images preserve identity to a degree approaching that of photographic stimuli.}
{To assess demographic diversity and ensure that intra-demographic similarity does not exceed inter-demographic similarity}, these findings confirm that our synthetic generation process successfully maintains identity consistency across neutral and expressive image pairs, making \synpain suitable for pairwise pain detection approaches that rely on comparing expressions to individual baselines.

Beyond identity consistency across expressions, we also examined identity diversity within demographic subgroups of \synpain. While the dataset contains 5,355 nominally unique identities, our analysis revealed systematic variations in effective identity diversity across demographic groups. We computed pairwise cosine similarities between all the identity vectors within each demographic subgroup and analyzed the proportion of highly similar pairs (cosine similarity$>$0.8) as an indicator of reduced diversity. Our analysis uncovered significant disparities in identity diversity, with notable differences both between genders and across ethnic/racial groups (\cref{tab:high_similarity}). 
{Chi-square tests revealed significant main effects for gender ($\chi^2(1) = 5188.77$, $p < 0.001$, Cramér's $V = 0.11$) and ethnicity ($\chi^2(4) = 1526.60$, $p < 0.001$, Cramér's $V = 0.06$), as well as a significant interaction between these factors ($\chi^2(9) = 6855.02$, $p < 0.001$, Cramér's $V = 0.12$).}
Women consistently exhibited lower identity diversity than men across all ethnicities/races, with 2.0-5.7\% of within-group pairs showing high similarity ($>$0.8) compared to only 0.3-1.9\% for men. Among women, East Asian women showed the lowest diversity (5.7\% of pairs above 0.8 similarity). \cref{fig:repeatedidentities} shows examples of East Asian and Middle Eastern female identities with the highest observed cosine similarity (0.94) between their facial encodings.

\begin{table}[h]
\centering
\caption{Percentage of within-group similarities above 0.8 by demographic group.\\{(ME: Middle Eastern, SA: South Asian, EA: East Asian)}}
\label{tab:high_similarity}
\begin{tabular}{lcccc}
\toprule
                & \textbf{Women (\%)} & \textbf{Men (\%)} & {\textbf{Diff}} & {\textbf{OR (W vs M)}} \\\midrule
\textbf{Black}  & 3.5 & 0.8 & {2.7} & {4.50 (3.98, 5.09)} \\
\textbf{White}  & 2.0 & 0.3 & {1.7} & {6.78 (5.56, 8.27)} \\
\textbf{ME}     & 4.7 & 0.3 & {4.4} & {16.43 (13.92, 19.39)} \\
\textbf{SA}     & 3.8 & 0.4 & {3.4} & {9.82 (8.36, 11.54)} \\
\textbf{EA}     & 5.7 & 1.9 & {3.8} & {3.12 (2.89, 3.38)} \\
\bottomrule
\end{tabular}
\end{table}

These findings suggest that while \synpain maintains strong identity consistency across expressions, the underlying generative process exhibits systematic bias in producing diverse identities across demographic groups. {One reason for the high percentage of within-group similarities among East Asian identities could be that we sampled from only nine nationalities to create prompts that generated these identities. In contrast, we used 43 nationalities for White identities, 21 for Black identities, 13 for South Asian identities, and 18 for Middle Eastern identities.} The observed pattern indicates potential limitations in the training data or generation algorithm that particularly affect women's facial diversity, with intersectional effects varying by ethnicity/race.

\begin{figure}[t]
    \centering
    \includegraphics[width=.2\columnwidth ]{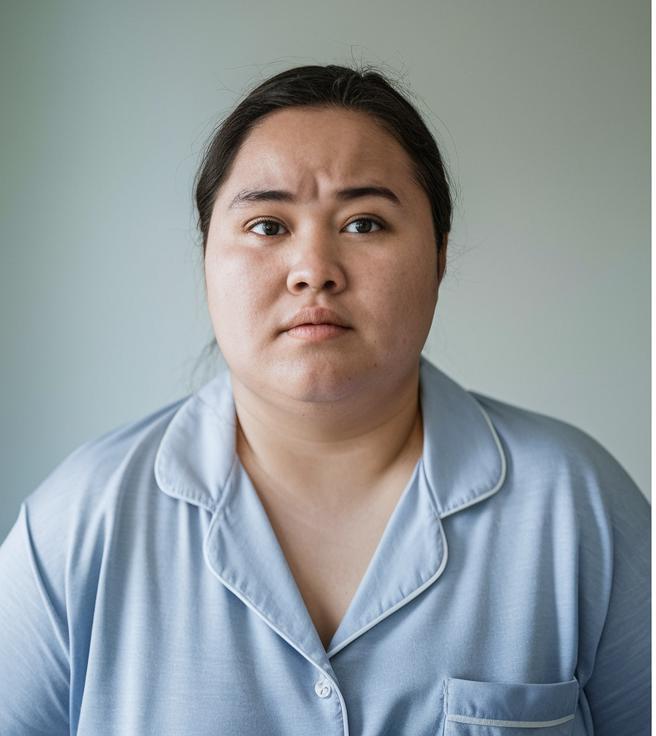} \hspace{-5pt}
    \includegraphics[width=.2\columnwidth ]{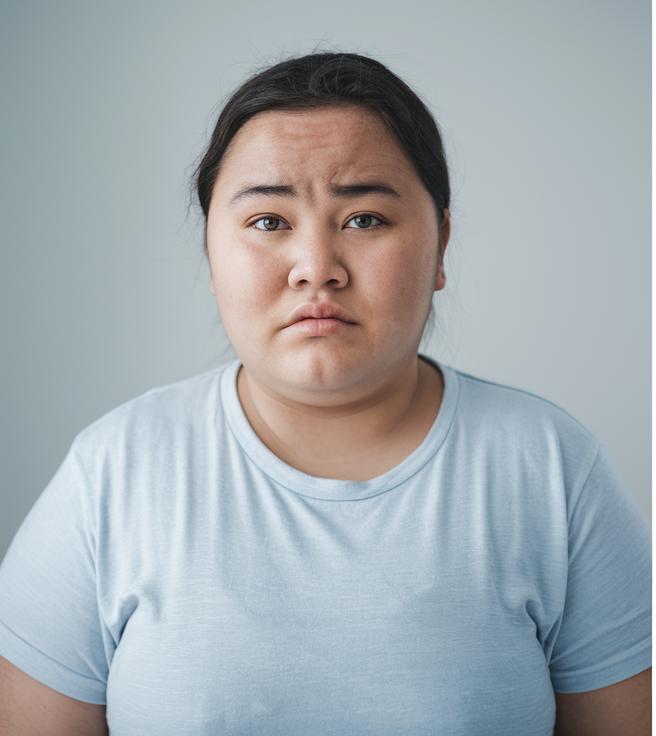} \hspace{5pt}
    \includegraphics[width=.2\columnwidth ]
    {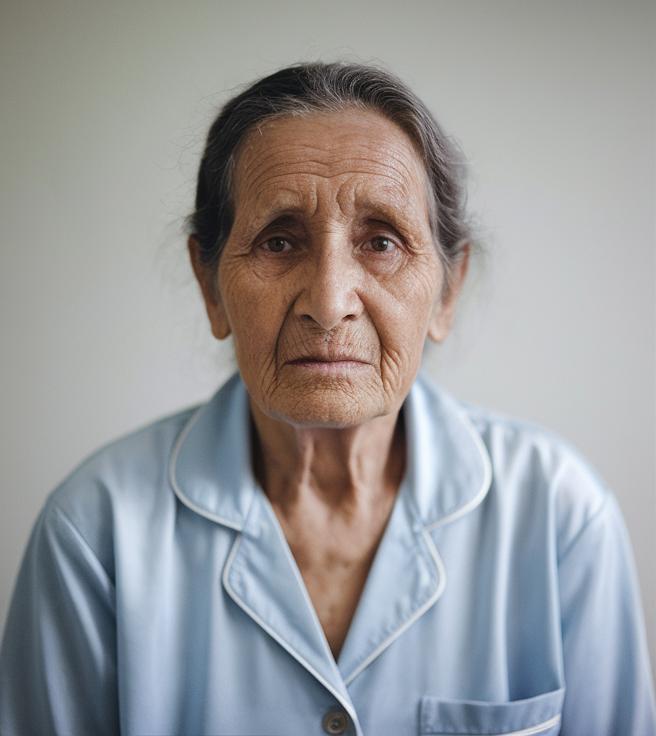} \hspace{-5pt}
    \includegraphics[width=.2\columnwidth ]{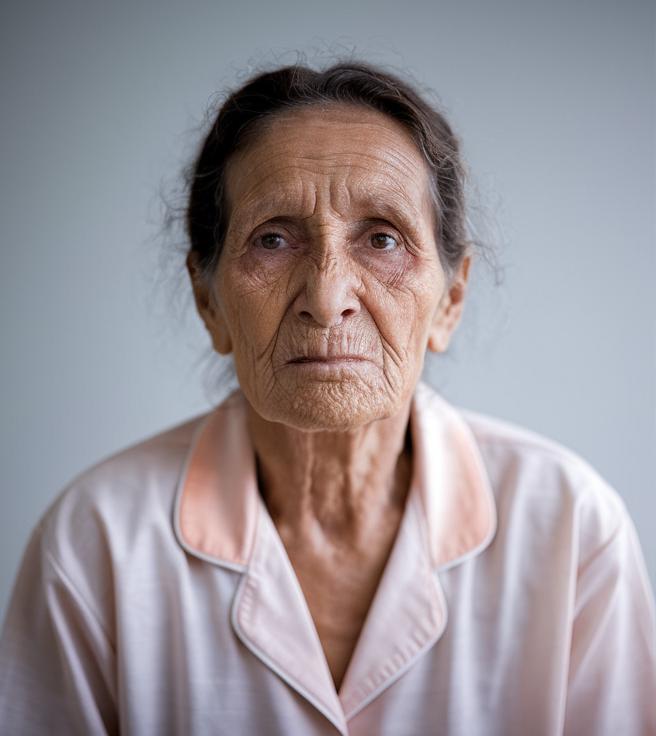}
    \caption{Examples of \synpain neutral images with very high (0.94) cosine similarity of their identities. Left: young East Asian females; Right: old Middle Eastern females. }
    \label{fig:repeatedidentities}
\end{figure}

\section{Within-Dataset Experiments}

Within-dataset experiments demonstrate the utility of the \synpain dataset for investigating algorithmic bias with respect to age, gender, and ethnicity/race. 
All experiments employ the pairwise pain detection model~\cite{rezaei2020unobtrusive}, trained from scratch with hyperparameters configured according to PainControl~\cite{PainControl}.

{All following tables report model performance as measured by three metrics: F1-score, average preceision (AP), and area under the receiver operating characteristic curve (AUROC). F1-score is the harmonic mean between precision and recall, which measure how well the model avoids false positives and false negatives, respectively. AP measures the area under the precision-recall curve, and is particularly useful when dealing with highly imbalanced datasets, e.g., if positive samples are rare. Lastly, AUROC measures the probability that the model will score a random positive sample higher than a random negative sample. Unlike AP, AUROC can be overly optimistic for highly imbalanced datasets, as it is easier to achieve high values without strong performance on the minority class~\cite{saito2015AUROC, davis2006AUROC}}.

\cref{tab:withindatasetgenderandage} shows pain detection results, quantified by AUROC, when the training set is mixed (5-fold cross-validation with 60/20/20 splits for train/validation/test) or stratified by age or gender. These results demonstrate that first, in 5-fold cross-validation, performance shows a substantial age bias favoring young faces over old faces (AUROC 0.755 vs. 0.692), but exhibits minimal gender bias with nearly equivalent performance between male and female faces (0.723 vs. 0.728). Second, age-stratified training reveals significant cross-age generalization challenges, with models trained exclusively on young faces showing a notable 5.7 percentage point drop when tested on older adults (0.752 vs. 0.695), while models trained only on older faces perform substantially worse overall, particularly struggling with older adult pain recognition (0.643 vs. 0.692 in mixed training). Third, gender stratification reveals a striking asymmetry in training data utility, where models trained exclusively on female data achieve strong performance across all demographics (0.711-0.798 range), while male-only training leads to severe performance degradation across all test groups (0.574-0.666 range), with the most dramatic decline occurring when testing on male faces themselves (0.574 vs. 0.723 in mixed training), suggesting that female facial expressions within \synpain provide more generalizable pain-related features than male expressions.

\cref{tab:withindatasetethnicity} shows pain detection results when the training set is stratified by ethnicity/race. These results clearly demonstrate significant variations in cross-ethnic/racial generalization capability, with South Asian training data showing the strongest cross-ethnic/racial performance (0.714-0.772 range across test ethnicities/races), while Black and East Asian training data exhibit the most limited generalization (Black: 0.631-0.652 range, East Asian: 0.572-0.717 range). This pattern aligns with the identity diversity analysis in \cref{tab:high_similarity}, where East Asian faces show the highest within-group similarity rates, potentially limiting the model's ability to learn generalizable pain-related features from this more homogeneous training set. Notably, all single-ethnicity/race training regimens underperform compared to mixed training, but the magnitude of this performance gap varies dramatically by ethnicity/race, with East Asian training showing particularly poor within-group performance (0.572 vs. 0.681 in mixed training), indicating substantial heterogeneity in the discriminative power of pain expressions across different ethnic/racial groups.

\begin{table}[h]
\caption{Pain detection results (AUROC) when the training is mixed (5-fold cross-validation) or stratified by age or gender.}
\label{tab:withindatasetgenderandage}
\centering

\begin{tabular}{llllll}
 \toprule
                        & \multicolumn{5}{c}{\textbf{Test}} \\\cmidrule{2-6}
\textbf{Training}       & \textbf{Young} & \textbf{Old} & \textbf{Man} & \textbf{Woman} & \textbf{All} \\ \midrule
\textbf{5-fold CV}      &  0.755   &  0.692  &  0.723 & 0.728 & 0.720 \\ \midrule
\textbf{Young}          &  0.752   &  0.695  &  0.722 & 0.725 & 0.718 \\ 
\textbf{Old}            &  0.707   &  0.643  &  0.676 & 0.731 & 0.698 \\ \midrule
\textbf{Male}           &  0.661   &  0.657  &  0.574 & 0.666 & 0.664 \\ 
\textbf{Female}         &  0.798   &  0.711  &  0.750 & 0.771 & 0.751 \\ 
\bottomrule
\end{tabular}
\end{table}

\begin{table}[h]
\caption{Pain detection results (AUROC) when the training is mixed (5-fold cross-validation) or stratified by ethnicity/race. \\(ME: Middle Eastern, SA: South Asian, EA: East Asian) }
\label{tab:withindatasetethnicity}
\centering
\begin{tabular}{lcccccc}
 \toprule
                        & \multicolumn{6}{c}{\textbf{Test}} \\\cmidrule{2-7}
\textbf{Training}       & \textbf{Black} & \textbf{White} & \textbf{ME} & \textbf{SA} & \textbf{EA} & \textbf{All} \\ \midrule
\textbf{5-fold CV}      &  0.714 &  0.727  &  0.732 & 0.757 & 0.681 & 0.720 \\ \midrule
\textbf{Black}          &  0.631   &  0.648  &  0.641 & 0.652 & 0.649 & 0.645 \\ 
\textbf{White}          &  0.718   &  0.734  &  0.736 & 0.739 & 0.608 & 0.697 \\ 
\textbf{Middle Eastern} &  0.705   &  0.746  &  0.752 & 0.751 & 0.639 & 0.709 \\ 
\textbf{South Asian}    &  0.731   &  0.771  &  0.768 & 0.772 & 0.714 & 0.747 \\ 
\textbf{East Asian}     &  0.664   &  0.717  &  0.693 & 0.703 & 0.572 & 0.685 \\ 
\bottomrule
\end{tabular}
\end{table}

\cref{tab:withindatasetleaveoneethnicityout} shows leave-one-ethnicity/race-out cross-validation results, when pain detection is quantified by the F1-score, AP, and AUROC. These results show that models exhibit substantial variation in cross-ethnic/racial generalization capability, with East Asian faces presenting the greatest challenge for pain detection when excluded from the training data. In contrast, Middle Eastern faces show the best cross-ethnic/racial generalization performance, followed closely by South Asian faces. The performance gap is substantial, spanning 8.4 percentage points in AUROC and 12.6 percentage points in AP, indicating significant disparities in how well pain expressions transfer across ethnic/racial boundaries.

\newcommand*{\MinNumber}{0.0}%
\newcommand*{\MidNumber}{0.5} %
\newcommand*{\MaxNumber}{1.0}%

\newcommand{\ApplyGradient}[1]{%
        \ifdim #1 pt > \MidNumber pt
            \pgfmathsetmacro{\PercentColor}{max(min(100.0*(#1 - \MidNumber)/(\MaxNumber-\MidNumber),100.0),0.00)} %
            \hspace{-0.33em}\colorbox{green!\PercentColor!yellow}{#1}
        \else
            \pgfmathsetmacro{\PercentColor}{max(min(100.0*(\MidNumber - #1)/(\MidNumber-\MinNumber),100.0),0.00)} %
            \hspace{-0.33em}\colorbox{red!\PercentColor!yellow}{#1}
        \fi
}

\begin{table}[h]
\caption{Leave-one-ethnicity/race-out cross-validation results.}
\label{tab:withindatasetleaveoneethnicityout}
\centering
\begin{tabular}{lccc}
\toprule
\textbf{Left-out set}       & \textbf{AUROC} & \textbf{AP} & \textbf{F1-score} \\ \midrule
\textbf{Black}          &  0.684   &  0.702  &  0.719 \\ 
\textbf{White}          &  0.706   &  0.697  &  0.715 \\ 
\textbf{Middle Eastern} &  0.717   &  0.728  &  0.714 \\ 
\textbf{South Asian}    &  0.711   &  0.724  &  0.693 \\ 
\textbf{East Asian}     &  0.633   &  0.602  &  0.666 \\ 
\bottomrule

\end{tabular}
\end{table}

\section{Evaluation of Pretrained Models}

Pretrained models can be evaluated on \synpain to examine their performance across gender, ethnicity/race, and age demographics. This evaluation is useful for identifying potential algorithmic biases as a prerequisite for developing mitigation strategies. 

The PwCT model~\cite{rezaei2020unobtrusive}, for instance, reported balanced results with respect to gender but was unable to assess performance across age and ethnicity/race due to the UofR dataset limitations. Here, we use \synpain to evaluate PwCT's performance on young versus old adults, men versus women, and across different ethnic/racial groups. The released PwCT model provides two checkpoints: one trained on UNBC-McMaster alone and another trained on UNBC-McMaster + UofR. All experiments in this section use the model trained on UNBC-McMaster + UofR, as it achieved superior performance compared to the UNBC-McMaster-only model across all demographic groups in \synpain.

\cref{tab:pairwiseonsynpain1} presents AUROC, AP, and F1-score of the PwCT model when evaluated on different demographic subgroups within \synpain. \cref{tab:pairwiseonsynpain2_auroc} further breaks down the AUROC results for specific demographic combinations, such as performance among South Asian women, young men, etc. 

The results reveal significant algorithmic biases in the pretrained PwCT model. \cref{tab:pairwiseonsynpain1} shows substantially worse performance on men compared to women (AUROC: 0.670 vs. 0.749). While the original PwCT paper~\cite{rezaei2020unobtrusive} reported relatively balanced gender performance with Pearson correlations of 0.46 for male faces and 0.50 for female faces when regressing to the pain levels, our evaluation reveals a much larger performance gap. This discrepancy may reflect our larger evaluation set of 5,355 distinct synthetic identities compared to the smaller cohort used in the previous study, which could reveal biases that were not detectable in smaller samples. The model also demonstrates substantial age bias, performing considerably worse on older faces compared to young adults (AUROC: 0.663 vs. 0.729).

Analysis of demographic intersections in \cref{tab:pairwiseonsynpain2_auroc} reveals additional disparities. Among older adults, performance varies considerably across ethnicities/races, with East Asian older faces showing substantially worse performance (AUROC: 0.623) compared to other ethnic/racial groups. This finding is unsurprising given that older East Asian individuals were underrepresented in the PwCT training datasets (UNBC-McMaster and UofR). These results demonstrate \synpain's value in uncovering algorithmic biases that may not be apparent when evaluating on smaller or less diverse datasets, highlighting the need for more inclusive training data in pain detection systems.

\begin{table}[h]
\caption{Pretrained PwCT model performance on different demographic subsets of \synpain.}
\label{tab:pairwiseonsynpain1}
\centering
\begin{tabular}{lccc}
\toprule
\textbf{Test set}           & \textbf{AUROC} & \textbf{AP}  & \textbf{F1-score} \\ \midrule
\textbf{\synpain}           &	0.696	&	0.710	&	0.696\\\midrule
\textbf{Young}	            &	0.729	&	0.713	&	0.681\\
\textbf{Old}                &	0.663	&	0.718	&	0.707\\\midrule
\textbf{Man}                &	0.670	&	0.693	&	0.675\\
\textbf{Woman}              &	0.749	&	0.767	&	0.720\\\midrule
\textbf{Black}              &	0.697	&	0.743	&	0.712\\
\textbf{White}              &	0.690	&	0.704	&	0.703\\
\textbf{Middle Eastern}     &	0.721	&	0.729	&	0.719\\
\textbf{South Asian}        &	0.667	&	0.686	&	0.664\\
\textbf{East Asian}         &	0.708	&	0.691	&	0.693\\
\bottomrule
\end{tabular}
\end{table}

\begin{table}[h]
\caption{AUROC breakdown for pretrained PwCT model across demographic combinations in \synpain.}
\label{tab:pairwiseonsynpain2_auroc}
\centering
\begin{tabular}{lccccc}
\toprule
\textbf{Test set}       & \textbf{Young} & \textbf{Old} & \textbf{Man} & \textbf{Woman} & \textbf{All}\\ \midrule
\textbf{\synpain} & 0.729 & 0.663 & 0.670 & 0.749 & 0.696 \\ \midrule
\textbf{Young}   & 0.729 & - & 0.688 & 0.795 & 0.729 \\ 
\textbf{Old}     & - & 0.663 & 0.647 & 0.714 & 0.663 \\ \midrule
\textbf{Black}   & 0.718 & 0.671 & 0.661 & 0.778 & 0.697 \\ 
\textbf{White}   & 0.740 & 0.656 & 0.682 & 0.726 & 0.690 \\ 
\textbf{Middle Eastern}      & 0.750 & 0.692 & 0.722 & 0.766 & 0.721 \\ 
\textbf{South Asian}         & 0.654 & 0.690 & 0.639 & 0.700 & 0.667 \\ 
\textbf{East Asian}         & 0.779 & 0.623 & 0.651 & 0.788 & 0.708       \\ 
\bottomrule
\end{tabular}
\end{table}

The \synpain dataset includes 40 synthetic videos transitioning from neutral to pain expressions, enabling qualitative evaluation of pain detection models. \cref{fig:SampleVideoFrames} shows a sample video sequence with frame-by-frame pain scores estimated by the PwCT model. The estimated PSPI scores demonstrate the model's ability to track the progression from neutral baseline to variations in pain expression over time. Note that while the PSPI calculated from the AUs provides integer values in the [0,16] range, the model outputs real numbers within this range.

\section{External Evaluation}

A critical application for synthetic data, and \synpain specifically, is training set augmentation to improve model performance on real-world data. In this section, we evaluate whether adding \synpain to the training data enhances PwCT model performance on the UofR dataset. {As mentioned in Section II.C, the UofR dataset is the only dataset of older adults with and without dementia and contains annotated data from 95 participants (74 women and 21 men), making it a perfect candidate dataset for the external evaluation of \synpain}.

We compare two training configurations: (1) Real-only training using UNBC-McMaster and UofR training folds via 5-fold cross-validation, and (2) Augmented training that adds the 2,895 old identities of the \synpain dataset to the real training data. We use only the older synthetic faces because the UofR dataset exclusively contains older adults, so we added age-matched synthetic data to maximize relevance and avoid potential domain mismatch issues that could arise from including younger synthetic faces. {The number of real samples (i.e., from UofR and UNBC-McMaster) amount to 225,834 images.}

\cref{tab:externalvalidation_auroc} and \cref{tab:externalvalidation_ap} present the results in terms of AUROC and AP, respectively. The real-only baseline replicates the performance reported in the original PwCT paper~\cite{rezaei2020unobtrusive} and subsequent work~\cite{PainControl}. The results demonstrate mixed but promising effects of synthetic data augmentation. While AUROC shows minimal overall change (0.775 to 0.778), there are notable improvements for the healthy older adult group (0.763 to 0.779, a 1.6 percentage point gain). {To assess statistical significance, we conducted a DeLong's test to compare the AUROC values and a paired t-test to compare the AP scores. The DeLong's test yielded a p-value of 0.081, indicating no statistically significant difference between the AUROC values. However, the paired t-test produced a p-value of 0.024, demonstrating a statistically significant improvement in AP scores.} More substantially, AP shows consistent improvements across all groups, with particularly notable gains in the healthy population (0.293 to 0.319, a {2.6 percentage point gain}) and overall performance (0.345 to 0.369, a {2.4 percentage point gain}). {As stated previously, AP better evaluates model performance on imbalanced datasets. The UofR test set exemplifies this, comprising only 21,769 positive samples (i.e., faces showing pain) out of 174,555 total samples (12.5\% positives)}. These results suggest that \synpain augmentation is particularly beneficial for improving precision in pain detection, which is clinically important for reducing false positive pain alerts in automated monitoring systems.

\begin{table}[h]
\centering
\caption{AUROC comparison of PwCT model performance on UofR test set with and without \synpain augmentation.}
\begin{tabular}{lccc}
\toprule
\textbf{Training Set} & \textbf{Dementia} & \textbf{Healthy} & \textbf{All} \\
\midrule
\textbf{Real-Only}  & 0.787 & 0.763 & 0.775 \\
\textbf{Real + \synpain Old} & 0.778 & 0.779 & 0.778 \\
\bottomrule
\end{tabular}
\label{tab:externalvalidation_auroc}
\end{table}

\begin{table}[h]
\centering
\caption{AP comparison of PwCT model performance on UofR test set with and without \ augmentation.}
\begin{tabular}{lccc}
\toprule
\textbf{Training Set} & \textbf{Dementia} & \textbf{Healthy} & \textbf{All} \\
\midrule
\textbf{Real-Only}  & 0.396 & 0.293 & 0.345 \\
\textbf{Real + \synpain Old } & 0.419 & 0.319 & 0.369 \\
\bottomrule
\end{tabular}
\label{tab:externalvalidation_ap}
\end{table}

\section{Discussion}

This work presents \synpain, a large-scale synthetic dataset of pain and non-pain facial expressions that addresses critical gaps in automated pain assessment research. The dataset enables systematic evaluation of algorithmic bias across age, gender, and ethnicity/race~--~demographics that are often underrepresented in existing pain datasets. Our findings confirm that synthetic facial expression data can effectively supplement real training data, addressing the persistent challenge of limited annotated datasets in specialized populations such as older adults with cognitive impairment.

Beyond the experimental validation and the publicly available dataset, this paper demonstrates the utility of synthetic data for measuring algorithmic bias and augmenting training datasets for improved model performance.

A key finding is that entirely synthetic images, generated solely through text prompts, prove effective for data augmentation, contrasting with prior work~\cite{PainControl} that showed expression transfer methods were not beneficial. This distinction is important because synthetic generation creates genuinely novel expressions rather than recycling existing pain patterns from limited real datasets.

The practical implications are significant: the entire \synpain dataset was generated using commercial generative AI tools for less than \$800 (Ideogram API and Runway standard annual subscription). This cost-effectiveness opens opportunities for researchers in related domains (such as neonatal pain assessment, orofacial evaluation in Bell's palsy, or other specialized clinical applications) to create tailored synthetic datasets with minimal financial barriers. {Beyond cost, \synpain also addresses several ethical concerns inherent to real-world datasets. Because all faces are synthetically generated, the dataset eliminates privacy risks associated with collecting and distributing images of real patients, which is a particularly sensitive concern in clinical contexts. Furthermore, by enabling deliberate control over demographic composition, synthetic generation offers a pathway to mitigate the demographic imbalances that have historically amplified bias in pain detection models, allowing researchers to construct datasets that are representative by design rather than by circumstance.}

While \synpain addresses many existing dataset limitations, several constraints remain: (1) A small percentage of identities within demographic groups show high similarity (cosine~similarity~$>$~0.8), with rates up to 5.7\% among certain groups, particularly affecting female synthetic faces more than male faces; (2) Although visual inspection was performed, approximately 1.2\% of images exhibit non-frontal poses (pitch~or~yaw~$>$~20\degree) or generation artifacts due to synthesis variability; (3) Categorizing identities into discrete ethnic/racial groups oversimplifies the complexity of human ethnicity and race, as many individuals have mixed heritage or may not clearly fit into predefined categories. This limitation reflects broader challenges in demographic classification and may not fully capture the continuous spectrum of human ethnic/racial diversity that exists in real-world populations.

\section{Conclusions and Future Work}
This work presents \synpain, the first publicly available, demographically diverse synthetic dataset for older adult pain detection. Our findings confirm that cost-effective synthetic expressions can improve real-world pain detection performance while revealing significant demographic disparities previously undetectable with smaller datasets. Future research directions include expanding demographic representation, generating longer video sequences {for extended temporal dynamics}, exploring domain adaptation techniques, incorporating lighting and pose variations to improve robustness to real-world imaging conditions, {and including multimodal signals (e.g., speech, physiological data)} to further improve real-world performance when augmenting with synthetic data. Moreover, there are many facial features that are affected by various diseases (e.g., stroke, Parkinson's disease, and Bell's palsy), which may contribute to misclassifications. Therefore, future work includes adding synthetic samples to represent subjects with these diseases to improve the model performance in the real world. Additionally, investigating how different prompt types (e.g., "in pain" vs. "lowered brow" vs. "groaning") affect validity, detection quality, and model performance represents an important avenue for optimization. Furthermore, examining the perceived intensity of generated pain expressions and how intensity variations interact with demographic features and perception biases could provide crucial insights for creating more effective training data. {Future dataset expansions could also systematically vary environmental factors such as lighting conditions and camera viewpoints, as clinical and real-world settings often differ substantially from the controlled frontal-face captures typical of existing pain datasets. Training on such variation may enhance model generalization across deployment contexts. Future work will further strengthen the validity of \synpain through expert evaluations and perceptual studies, confirming that synthetic expressions faithfully capture clinically meaningful facial pain indicators.}

\section*{Acknowledgements}

This research was made possible with funding and support from AGE-WELL, the Canadian Institutes of Health Research (CIHR), AMS Healthcare, the Natural Sciences and Engineering Research Council of Canada (NSERC), the Data Science Institute (DSI) at the University of Toronto, and the KITE Research Institute, Toronto Rehabilitation Institute, UHN.

\bibliography{SynPain}

@article{achterberg2021chronic,
  title={Are chronic pain patients with dementia being undermedicated?},
  author={Achterberg, Wilco P and Erdal, Ane and Husebo, Bettina S and Kunz, Miriam and Lautenbacher, Stefan},
  journal={Journal of Pain Research},
  pages={431--439},
  year={2021},
  publisher={Taylor \& Francis}
}

@article{cipher2006behavioral,
  title={Behavioral manifestations of pain in the demented elderly},
  author={Cipher, Daisha J and Clifford, P Andrew and Roper, Kristi D},
  journal={Journal of the American Medical Directors Association},
  volume={7},
  number={6},
  pages={355--365},
  year={2006},
  publisher={Elsevier}
}

@article{avent2013establishing,
  title={Establishing the motivations of patients with dementia and cognitive impairment and their carers in joining a dementia research register (DemReg)},
  author={Avent, Cerian and Curry, Lisa and Gregory, Sarah and Marquardt, Sonia and Pae, Lauren and Wilson, Danielle and Ritchie, Karen and Ritchie, Craig W},
  journal={International psychogeriatrics},
  volume={25},
  number={6},
  pages={963--971},
  year={2013},
  publisher={Cambridge University Press}
}

@inproceedings{buolamwini2018gender,
  title={Gender shades: Intersectional accuracy disparities in commercial gender classification},
  author={Buolamwini, Joy and Gebru, Timnit},
  booktitle={Conference on fairness, accountability and transparency},
  pages={77--91},
  year={2018},
  organization={PMLR}
}

@article{rezaei2020unobtrusive,
  title={Unobtrusive pain monitoring in older adults with dementia using pairwise and contrastive training},
  author={Rezaei, Siavash and Moturu, Abhishek and Zhao, Shun and Prkachin, Kenneth M and Hadjistavropoulos, Thomas and Taati, Babak},
  journal={IEEE Journal of Biomedical and Health Informatics},
  volume={25},
  number={5},
  pages={1450--1462},
  year={2020},
  publisher={IEEE}
}

@article{prkachin2008structure,
  title={The structure, reliability and validity of pain expression: Evidence from patients with shoulder pain},
  author={Prkachin, Kenneth M and Solomon, Patricia E},
  journal={Pain},
  volume={139},
  number={2},
  pages={267--274},
  year={2008},
  publisher={Elsevier}
}

@article{chan2014evidence,
  title={Evidence-based development and initial validation of the pain assessment checklist for seniors with limited ability to communicate-{II} ({PACSLAC-II})},
  author={Chan, Sarah and Hadjistavropoulos, Thomas and Williams, Jaime and Lints-Martindale, Amanda},
  journal={The Clinical journal of pain},
  volume={30},
  number={9},
  pages={816--824},
  year={2014},
  publisher={LWW}
}

@article{ekman1978facial,
  title={Facial action coding system},
  author={Ekman, Paul and Friesen, Wallace V},
  journal={Environmental Psychology \& Nonverbal Behavior},
  year={1978}
}

@inproceedings{lucey2011painful,
  title={Painful data: The {UNBC-McMaster} shoulder pain expression archive database},
  author={Lucey, Patrick and Cohn, Jeffrey F and Prkachin, Kenneth M and Solomon, Patricia E and Matthews, Iain},
  booktitle={2011 IEEE International Conference on Automatic Face \& Gesture Recognition (FG)},
  pages={57--64},
  year={2011},
  organization={IEEE}
}

@inproceedings{walter2013biovid,
  title={The {BioVid} heat pain database data for the advancement and systematic validation of an automated pain recognition system},
  author={Walter, Steffen and Gruss, Sascha and Ehleiter, Hagen and Tan, Junwen and Traue, Harald C and Werner, Philipp and Al-Hamadi, Ayoub and Crawcour, Stephen and Andrade, Adriano O and da Silva, Gustavo Moreira},
  booktitle={2013 IEEE international conference on cybernetics (CYBCO)},
  pages={128--131},
  year={2013},
  organization={IEEE}
}

@article{bandini2020new,
  title={A new dataset for facial motion analysis in individuals with neurological disorders},
  author={Bandini, Andrea and Rezaei, Sia and Guar{\'\i}n, Diego L and Kulkarni, Madhura and Lim, Derrick and Boulos, Mark I and Zinman, Lorne and Yunusova, Yana and Taati, Babak},
  journal={IEEE Journal of Biomedical and Health Informatics},
  volume={25},
  number={4},
  pages={1111--1119},
  year={2020},
  publisher={IEEE}
}

@inproceedings{asgarian2019limitations,
  title={Limitations and Biases in Facial Landmark Detection D An Empirical Study on Older Adults with Dementia.},
  author={Asgarian, Azin and Zhao, Shun and Ashraf, Ahmed Bilal and Browne, M Erin and Prkachin, Kenneth M and Mihailidis, Alex and Hadjistavropoulos, Thomas and Taati, Babak},
  booktitle={CVPR workshops},
  pages={28--36},
  year={2019}
}

@article{gkikas2023automatic,
  title={Automatic assessment of pain based on deep learning methods: A systematic review},
  author={Gkikas, Stefanos and Tsiknakis, Manolis},
  journal={Computer methods and programs in biomedicine},
  volume={231},
  pages={107365},
  year={2023},
  publisher={Elsevier}
}

@article{mavadati2013disfa,
  title={{DISFA}: A spontaneous facial action intensity database},
  author={Mavadati, S Mohammad and Mahoor, Mohammad H and Bartlett, Kevin and Trinh, Philip and Cohn, Jeffrey F},
  journal={IEEE Transactions on Affective Computing},
  volume={4},
  number={2},
  pages={151--160},
  year={2013},
  publisher={IEEE}
}

@article{zhang2014bp4d,
  title={{BP4D}-spontaneous: a high-resolution spontaneous 3d dynamic facial expression database},
  author={Zhang, Xing and Yin, Lijun and Cohn, Jeffrey F and Canavan, Shaun and Reale, Michael and Horowitz, Andy and Liu, Peng and Girard, Jeffrey M},
  journal={Image and Vision Computing},
  volume={32},
  number={10},
  pages={692--706},
  year={2014},
  publisher={Elsevier}
}

@inproceedings{zhang2016multimodal,
  title={Multimodal spontaneous emotion corpus for human behavior analysis},
  author={Zhang, Zheng and Girard, Jeff M and Wu, Yue and Zhang, Xing and Liu, Peng and Ciftci, Umur and Canavan, Shaun and Reale, Michael and Horowitz, Andy and Yang, Huiyuan and others},
  booktitle={Proceedings of the IEEE conference on computer vision and pattern recognition},
  pages={3438--3446},
  year={2016}
}

@inproceedings{savran2008bosphorus,
  title={Bosphorus database for 3D face analysis},
  author={Savran, Arman and Aly{\"u}z, Ne{\c{s}}e and Dibeklio{\u{g}}lu, Hamdi and {\c{C}}eliktutan, Oya and G{\"o}kberk, Berk and Sankur, B{\"u}lent and Akarun, Lale},
  booktitle={Biometrics and Identity Management: First European Workshop, BIOID 2008, Roskilde, Denmark, May 7-9, 2008. Revised Selected Papers 1},
  pages={47--56},
  year={2008},
  organization={Springer}
}

@article{egbujie2024trajectories,
  title={Trajectories of functional decline and predictors in long-term care settings: a retrospective cohort analysis of Canadian nursing home residents},
  author={Egbujie, Bonaventure Amandi and Turcotte, Luke Andrew and Heckman, George and Hirdes, John P},
  journal={Age and Ageing},
  volume={53},
  number={12},
  pages={afae264},
  year={2024},
  publisher={Oxford University Press}
}

@inproceedings{lucey2010extended,
  title={The extended cohn-kanade dataset ({CK+}): A complete dataset for action unit and emotion-specified expression},
  author={Lucey, Patrick and Cohn, Jeffrey F and Kanade, Takeo and Saragih, Jason and Ambadar, Zara and Matthews, Iain},
  booktitle={2010 ieee computer society conference on computer vision and pattern recognition-workshops},
  pages={94--101},
  year={2010},
  organization={IEEE}
}

@article{stopyn2025real,
  title={Real-time evaluation of an automated computer vision system to monitor pain behavior in older adults},
  author={Stopyn, Rhonda JN and Moturu, Abhishek and Taati, Babak and Hadjistavropoulos, Thomas},
  journal={Journal of Rehabilitation and Assistive Technologies Engineering},
  volume={12},
  pages={20556683251313762},
  year={2025},
  publisher={SAGE Publications Sage UK: London, England}
}

@article{brahnam2006machine,
  title={Machine recognition and representation of neonatal facial displays of acute pain},
  author={Brahnam, Sheryl and Chuang, Chao-Fa and Shih, Frank Y and Slack, Melinda R},
  journal={Artificial intelligence in medicine},
  volume={36},
  number={3},
  pages={211--222},
  year={2006},
  publisher={Elsevier}
}

@article{ma2105cfd,
    author={Ma, Debbie and Correll, Joshua and Wittenbrink, Bernd},
    title={The Chicago face database: A free stimulus set of faces and norming data},
    journal={Behavior Research Methods},
    volume={47},
    number={4},
    pages={1122--1135},
    year={2015},
    publisher={Psychonomic Society, Inc.}
}

@inproceedings{paraperas2024arc2face,
      title={Arc2Face: A Foundation Model for ID-Consistent Human Faces},
      author={Paraperas Papantoniou, Foivos and Lattas, Alexandros and Moschoglou, Stylianos and Deng, Jiankang and Kainz, Bernhard and Zafeiriou, Stefanos},
      booktitle={Proceedings of the European Conference on Computer Vision (ECCV)},
      year={2024}
}

@article{wang2024magicface,
      title={MagicFace: Training-free Universal-Style Human Image Customized Synthesis}, 
      author={Wang, Yibin and Zhang, Weizhong and Jin, Cheng},
      journal={arXiv preprint arXiv:2408.07433},
      year={2024}
}

@misc{mishima2025facecrafter,
      title={FaceCrafter: Identity-Conditional Diffusion with Disentangled Control over Facial Pose, Expression, and Emotion}, 
      author={Kazuaki Mishima and Antoni Bigata Casademunt and Stavros Petridis and Maja Pantic and Kenji Suzuki},
      year={2025},
      eprint={2505.15313},
      archivePrefix={arXiv},
      primaryClass={cs.CV},
      url={https://arxiv.org/abs/2505.15313}, 
}

@article{saito2015AUROC,
  title={The Precision-Recall Plot Is More Informative than the ROC Plot When Evaluating Binary Classifiers on Imbalanced Datasets},
  author={Saito, Takaya and Rehmsmeier, Marc},
  journal={PLoS ONE},
  volume={10},
  number={3},
  year={2015}
}

@inproceedings{davis2006AUROC,
    author={Davis, Jesse and Goadrich, Mark},
    title={The relationship between Precision-Recall and ROC curves},
    booktitle={ICML '06: Proceedings of the 23rd International Conference on Machine Learning},
    year={2006} 
}

@inproceedings{zarghami2023pain,
  title={Pain detection in masked faces during procedural sedation},
  author={Zarghami, Y and Mafeld, S and Conway, A and Taati, Babak},
  booktitle={2023 IEEE 17th International Conference on Automatic Face and Gesture Recognition (FG)},
  pages={1--6},
  year={2023},
  organization={IEEE}
}

@article{kunz2017problems,
  title={Problems of video-based pain detection in patients with dementia: a road map to an interdisciplinary solution},
  author={Kunz, Miriam and Seuss, Dominik and Hassan, Teena and Garbas, Jens U and Siebers, Michael and Schmid, Ute and Sch{\"o}berl, Michael and Lautenbacher, Stefan},
  journal={BMC geriatrics},
  volume={17},
  pages={1--8},
  year={2017},
  publisher={Springer}
}

@inproceedings{ashraf2007painful,
  title={The painful face: Pain expression recognition using active appearance models},
  author={Ashraf, Ahmed Bilal and Lucey, Simon and Cohn, Jeffrey F and Chen, Tsuhan and Ambadar, Zara and Prkachin, Ken and Solomon, Patty and Theobald, Barry J},
  booktitle={Proceedings of the 9th international conference on Multimodal interfaces},
  pages={9--14},
  year={2007}
}

@article{hadjistavropoulos2007interdisciplinary,
  title={An interdisciplinary expert consensus statement on assessment of pain in older persons},
  author={Hadjistavropoulos, Thomas and Herr, Keela and Turk, Dennis C and Fine, Perry G and Dworkin, Robert H and Helme, Robert and Jackson, Kenneth and Parmelee, Patricia A and Rudy, Thomas E and Beattie, B Lynn and others},
  journal={Clinical {Journal} of {Pain}},
  volume={23},
  pages={S1--S43},
  year={2007},
  publisher={LWW}
}

@inproceedings{xu2019pain,
  title={Pain Evaluation in Video using Extended Multitask Learning from Multidimensional Measurements.},
  author={Xu, Xiaojing and Huang, Jeannie S and De Sa, Virginia R},
  booktitle={ML4H@ NeurIPS},
  pages={141--154},
  year={2019}
}

@article{tavakolian2019spatiotemporal,
  title={A spatiotemporal convolutional neural network for automatic pain intensity estimation from facial dynamics},
  author={Tavakolian, Mohammad and Hadid, Abdenour},
  journal={International Journal of Computer Vision},
  volume={127},
  pages={1413--1425},
  year={2019},
  publisher={Springer}
}

@inproceedings{rau2024video,
  title={Video Swin Transformers in Pain Detection: A Comprehensive Evaluation of Effectiveness, Generalizability, and Explainability},
  author={Rau, Maximilian and Ertugrul, Itir Onal},
  booktitle={2024 12th International Conference on Affective Computing and Intelligent Interaction Workshops and Demos (ACIIW)},
  pages={22--30},
  year={2024},
  organization={IEEE}
}

@article{fiorentini2022fully,
  title={Fully-attentive and interpretable: vision and video vision transformers for pain detection},
  author={Fiorentini, Giacomo and Ertugrul, Itir Onal and Salah, Albert Ali},
  journal={arXiv preprint arXiv:2210.15769},
  year={2022}
}

@inproceedings{yang2019facs3d,
  title={{FACS3D-Net}: {3D} convolution based spatiotemporal representation for action unit detection},
  author={Yang, Le and Ertugrul, Itir Onal and Cohn, Jeffrey F and Hammal, Zakia and Jiang, Dongmei and Sahli, Hichem},
  booktitle={2019 8th International conference on affective computing and intelligent interaction (ACII)},
  pages={538--544},
  year={2019},
  organization={IEEE}
}

@article{onal2019d,
  title={{D-PAttNet}: Dynamic patch-attentive deep network for action unit detection},
  author={Onal Ertugrul, Itir and Yang, Le and Jeni, L{\'a}szl{\'o} A and Cohn, Jeffrey F},
  journal={Frontiers in computer science},
  volume={1},
  pages={11},
  year={2019},
  publisher={Frontiers Media SA}
}

@inproceedings{yuan2024auformer,
  title={{AUFormer}: Vision transformers are parameter-efficient facial action unit detectors},
  author={Yuan, Kaishen and Yu, Zitong and Liu, Xin and Xie, Weicheng and Yue, Huanjing and Yang, Jingyu},
  booktitle={European Conference on Computer Vision},
  pages={427--445},
  year={2024},
  organization={Springer}
}

@article{ning2024representation,
  title={Representation learning and identity adversarial training for facial behavior understanding},
  author={Ning, Mang and Salah, Albert Ali and Ertugrul, Itir Onal},
  journal={arXiv preprint arXiv:2407.11243},
  year={2024}
}

@article{cheong2023py,
  title={{Py-Feat}: Python facial expression analysis toolbox},
  author={Cheong, Jin Hyun and Jolly, Eshin and Xie, Tiankang and Byrne, Sophie and Kenney, Matthew and Chang, Luke J},
  journal={Affective Science},
  volume={4},
  number={4},
  pages={781--796},
  year={2023},
  publisher={Springer}
}

@article{guarin2020toward,
  title={Toward an automatic system for computer-aided assessment in facial palsy},
  author={Guarin, Diego L and Yunusova, Yana and Taati, Babak and Dusseldorp, Joseph R and Mohan, Suresh and Tavares, Joana and van Veen, Martinus M and Fortier, Emily and Hadlock, Tessa A and Jowett, Nate},
  journal={Facial Plastic Surgery \& Aesthetic Medicine},
  volume={22},
  number={1},
  pages={42--49},
  year={2020},
  publisher={Mary Ann Liebert, Inc., publishers 140 Huguenot Street, 3rd Floor New~…}
}

@inproceedings{yaar2002fifty,
  title={Fifty years of skin aging},
  author={Yaar, Mina and Eller, Mark S and Gilchrest, Barbara A},
  booktitle={Journal of Investigative Dermatology Symposium Proceedings},
  volume={7},
  number={1},
  pages={51--58},
  year={2002},
  organization={Elsevier}
}

@article{quan2015role,
  title={Role of age-associated alterations of the dermal extracellular matrix microenvironment in human skin aging: a mini-review},
  author={Quan, Taihao and Fisher, Gary J},
  journal={Gerontology},
  volume={61},
  number={5},
  pages={427--434},
  year={2015},
  publisher={S. Karger AG}
}

@article{chung2001modulation,
  title={Modulation of skin collagen metabolism in aged and photoaged human skin in vivo},
  author={Chung, Jin Ho and Seo, Jin Young and Choi, Hai Ryung and Lee, Mi Kyung and Youn, Choon Shik and Rhie, Gi-eun and Cho, Kwang Hyun and Kim, Kyu Han and Park, Kyung Chan and Eun, Hee Chul},
  journal={Journal of Investigative Dermatology},
  volume={117},
  number={5},
  pages={1218--1224},
  year={2001},
  publisher={Elsevier}
}

@article{hadjistavropoulos2014pain,
  title={Pain assessment in elderly adults with dementia},
  author={Hadjistavropoulos, Thomas and Herr, Keela and Prkachin, Kenneth M and Craig, Kenneth D and Gibson, Stephen J and Lukas, Albert and Smith, Jonathan H},
  journal={The Lancet Neurology},
  volume={13},
  number={12},
  pages={1216--1227},
  year={2014},
  publisher={Elsevier}
}

@article{pringle2021pain,
  title={Pain assessment and management in care homes: understanding the context through a scoping review},
  author={Pringle, Jan and Mellado, Ana Sofia Alvarado V{\'a}zquez and Haraldsdottir, Erna and Kelly, Fiona and Hockley, Jo},
  journal={BMC geriatrics},
  volume={21},
  pages={1--13},
  year={2021},
  publisher={Springer}
}

@article{weissman1999pain,
  title={Pain assessment and management in the long-term care setting},
  author={Weissman, David E and Matson, Sandra},
  journal={Theoretical Medicine and Bioethics},
  volume={20},
  pages={31--43},
  year={1999},
  publisher={Springer}
}

@article{herr2019pain,
  title={Pain assessment in the patient unable to self-report: clinical practice recommendations in support of the ASPMN 2019 position statement},
  author={Herr, Keela and Coyne, Patrick J and Ely, Elizabeth and G{\'e}linas, C{\'e}line and Manworren, Renee CB},
  journal={Pain Management Nursing},
  volume={20},
  number={5},
  pages={404--417},
  year={2019},
  publisher={Elsevier}
}

@article{herr2011pain,
  title={Pain assessment in the patient unable to self-report: position statement with clinical practice recommendations},
  author={Herr, Keela and Coyne, Patrick J and McCaffery, Margo and Manworren, Renee and Merkel, Sandra},
  journal={Pain management nursing},
  volume={12},
  number={4},
  pages={230--250},
  year={2011},
  publisher={Elsevier}
}

@article{konecny2011facial,
  title={Facial paresis after stroke and its impact on patients' facial movement and mental status.},
  author={Konecny, Petr and Elfmark, Milan and Urbanek, Karel},
  journal={Journal of Rehabilitation Medicine},
  volume={43},
  number={1},
  pages={73--75},
  year={2011}
}

@article{taati2019algorithmic,
  title={Algorithmic bias in clinical populations—evaluating and improving facial analysis technology in older adults with dementia},
  author={Taati, Babak and Zhao, Shun and Ashraf, Ahmed B and Asgarian, Azin and Browne, M Erin and Prkachin, Kenneth M and Mihailidis, Alex and Hadjistavropoulos, Thomas},
  journal={IEEE access},
  volume={7},
  pages={25527--25534},
  year={2019},
  publisher={IEEE}
}

@article{volk2019facial,
  title={Facial motor and non-motor disabilities in patients with central facial paresis: a prospective cohort study},
  author={Volk, Gerd Fabian and Steinerstauch, Anika and Lorenz, Annegret and Modersohn, Luise and Mothes, Oliver and Denzler, Joachim and Klingner, Carsten M and Hamzei, Farsin and Guntinas-Lichius, Orlando},
  journal={Journal of neurology},
  volume={266},
  pages={46--56},
  year={2019},
  publisher={Springer}
}

@article{guliani2021pain,
  title={Pain-related health care costs for long-term care residents},
  author={Guliani, Harminder and Hadjistavropoulos, Thomas and Jin, Shan and Lix, Lisa M},
  journal={BMC geriatrics},
  volume={21},
  pages={1--14},
  year={2021},
  publisher={Springer}
}

@article{gagnon2013development,
  title={Development and Mixed-Methods Evaluation of a Pain Assessment Video Training Program for Long-Term Care Staff},
  author={Gagnon, Michelle M and Hadjistavropoulos, Thomas and Williams, Jaime},
  journal={Pain Research and Management},
  volume={18},
  number={6},
  pages={307--312},
  year={2013},
  publisher={Wiley Online Library}
}

@article{werner2019automatic,
  title={Automatic recognition methods supporting pain assessment: A survey},
  author={Werner, Philipp and Lopez-Martinez, Daniel and Walter, Steffen and Al-Hamadi, Ayoub and Gruss, Sascha and Picard, Rosalind W},
  journal={IEEE Transactions on Affective Computing},
  volume={13},
  number={1},
  pages={530--552},
  year={2019},
  publisher={IEEE}
}

@article{benavent2023comprehensive,
  title={A comprehensive study on pain assessment from multimodal sensor data},
  author={Benavent-Lledo, Manuel and Mulero-P{\'e}rez, David and Ortiz-Perez, David and Rodriguez-Juan, Javier and Berenguer-Agullo, Adrian and Psarrou, Alexandra and Garcia-Rodriguez, Jose},
  journal={Sensors},
  volume={23},
  number={24},
  pages={9675},
  year={2023},
  publisher={MDPI}
}

@article{kobayashi2021semi,
  title={Semi-automated tracking of pain in critical care patients using artificial intelligence: a retrospective observational study},
  author={Kobayashi, Naoya and Shiga, Takuya and Ikumi, Saori and Watanabe, Kazuki and Murakami, Hitoshi and Yamauchi, Masanori},
  journal={Scientific Reports},
  volume={11},
  number={1},
  pages={5229},
  year={2021},
  publisher={Nature Publishing Group UK London}
}

@article{zamzmi2019comprehensive,
  title={A comprehensive and context-sensitive neonatal pain assessment using computer vision},
  author={Zamzmi, Ghada and Pai, Chih-Yun and Goldgof, Dmitry and Kasturi, Rangachar and Ashmeade, Terri and Sun, Yu},
  journal={IEEE Transactions on Affective Computing},
  volume={13},
  number={1},
  pages={28--45},
  year={2019},
  publisher={IEEE}
}

@article{zamzmi2019convolutional,
  title={Convolutional neural networks for neonatal pain assessment},
  author={Zamzmi, Ghada and Paul, Rahul and Salekin, Md Sirajus and Goldgof, Dmitry and Kasturi, Rangachar and Ho, Thao and Sun, Yu},
  journal={IEEE Transactions on Biometrics, Behavior, and Identity Science},
  volume={1},
  number={3},
  pages={192--200},
  year={2019},
  publisher={IEEE}
}

@article{brahnam2023neonatal,
  title={Neonatal pain detection in videos using the iCOPEvid dataset and an ensemble of descriptors extracted from Gaussian of Local Descriptors},
  author={Brahnam, Sheryl and Nanni, Loris and McMurtrey, Shannon and Lumini, Alessandra and Brattin, Rick and Slack, Melinda and Barrier, Tonya},
  journal={Applied Computing and Informatics},
  volume={19},
  number={1/2},
  pages={122--143},
  year={2023},
  publisher={Emerald Publishing Limited}
}

@inproceedings{zhaox,
  title={{X-NeMo}: Expressive Neural Motion Reenactment via Disentangled Latent Attention},
  author={Zhao, Xiaochen and Xu, Hongyi and Song, Guoxian and Xie, You and Zhang, Chenxu and Li, Xiu and Luo, Linjie and Suo, Jinli and Liu, Yebin},
  booktitle={The Thirteenth International Conference on Learning Representations},
  year={2025}
}

@inproceedings{rochow2024fsrt,
  title={{FSRT}: Facial scene representation transformer for face reenactment from factorized appearance head-pose and facial expression features},
  author={Rochow, Andre and Schwarz, Max and Behnke, Sven},
  booktitle={Proceedings of the IEEE/CVF conference on computer vision and pattern recognition},
  pages={7716--7726},
  year={2024}
}

@article{guo2024liveportrait,
  title={{LivePortrait}: Efficient portrait animation with stitching and retargeting control},
  author={Guo, Jianzhu and Zhang, Dingyun and Liu, Xiaoqiang and Zhong, Zhizhou and Zhang, Yuan and Wan, Pengfei and Zhang, Di},
  journal={arXiv preprint arXiv:2407.03168},
  year={2024}
}

@article{siarohin2019first,
  title={First order motion model for image animation},
  author={Siarohin, Aliaksandr and Lathuili{\`e}re, St{\'e}phane and Tulyakov, Sergey and Ricci, Elisa and Sebe, Nicu},
  journal={Advances in neural information processing systems},
  volume={32},
  year={2019}
}

@inproceedings{kanade2000comprehensive,
  title={Comprehensive database for facial expression analysis},
  author={Kanade, Takeo and Cohn, Jeffrey F and Tian, Yingli},
  booktitle={Proceedings fourth IEEE international conference on automatic face and gesture recognition (cat. No. PR00580)},
  pages={46--53},
  year={2000},
  organization={IEEE}
}

@inproceedings{chen2024dsl,
  title={{DSL-FIQA}: Assessing facial image quality via dual-set degradation learning and landmark-guided transformer},
  author={Chen, Wei-Ting and Krishnan, Gurunandan and Gao, Qiang and Kuo, Sy-Yen and Ma, Sizhou and Wang, Jian},
  booktitle={Proceedings of the IEEE/CVF Conference on Computer Vision and Pattern Recognition},
  pages={2931--2941},
  year={2024}
}

@article{PainControl,
  title={PainControl: Identity-Preserving Pain Expression Transfer with Generative Diffusion Models},
  author={Yasamin Zarghami and Vida Adeli and Hailey Reimer and Thomas Hadjistavropoulos and Babak Taati},
  journal={Under Review},
  year={2025}
}

@incollection{hadjistavropoulos2005assessing,
  author = {Hadjistavropoulos, Thomas},
  title = {Assessing pain in older persons with severe limitations in ability to communicate},
  booktitle = {Pain in older persons},
  editor = {Gibson, S. and Weiner, D.},
  pages = {135--151},
  year = {2005},
  publisher = {IASP Press},
  address = {Seattle}
}

@data{SP3/WCXMAP_2025,
author = {Taati, Babak and Muzammil, Muhammad and Zarghami, Yasamin and Moturu, Abhishek and Kazerouni, Amirhossein and Reimer, Hailey and Mihailidis, Alex and Hadjistavropoulos, Thomas},
publisher = {Borealis},
title = {{\synpain Data Repository}},
year = {2025},
version = {V1},
doi = {10.5683/SP3/WCXMAP},
url = {https://doi.org/10.5683/SP3/WCXMAP}
}
\bibliographystyle{IEEEtran}

\end{document}